\documentclass[journal]{IEEEtran}
\usepackage[T1]{fontenc}
\usepackage{amsmath,amssymb,amsfonts}
\usepackage{graphicx}
\usepackage{booktabs}
\usepackage{array}
\usepackage{multirow}
\usepackage{xcolor}
\usepackage{algorithm}
\usepackage{algpseudocode}
\renewcommand{\algorithmicrequire}{\textbf{Input:}}
\usepackage[nocompress]{cite}
\usepackage[hidelinks]{hyperref}
\usepackage[capitalise,nameinlink]{cleveref}

\crefname{section}{Section}{Sections}
\crefname{subsection}{Section}{Sections}
\crefname{subsubsection}{Section}{Sections}
\crefname{table}{Table}{Tables}
\crefname{figure}{Fig.}{Figs.}
\definecolor{figref}{HTML}{1F5FAD}
\crefformat{figure}{{\color{figref}#2Fig.~#1#3}}
\Crefformat{figure}{{\color{figref}#2Fig.~#1#3}}
\crefname{algorithm}{Algorithm}{Algorithms}
\crefname{equation}{}{}
\creflabelformat{equation}{(#2#1#3)}

\begin{document}

\title{From Weights to Features: Sparse-Autoencoder-Guided Activation Regularization for Continual Learning in Large Language Models}

\author{Evan~Ning,
        Wei~Xue,
        Dong~Lou,
        and~Yike~Guo%
\thanks{The authors are with The Hong Kong University of Science and Technology, Hong Kong SAR, China (e-mail: ekning@connect.ust.hk).}%
\thanks{Corresponding author: Evan Ning.}}

\markboth{IEEE Transactions on Audio, Speech and Language Processing}%
{Ning \MakeLowercase{\textit{et al.}}: SAE-Guided Activation Regularization for Continual Learning}

\maketitle

\begin{abstract}
Weight-protecting regularizers such as elastic weight consolidation (EWC) are a standard defense against catastrophic forgetting; however, in the sequential fine-tuning of large language models, they exhibit a stability--plasticity tradeoff that penalty tuning cannot resolve. We attribute this limitation in part to polysemanticity: because each weight participates in many unrelated concepts, a per-weight importance score has difficulty isolating the knowledge that requires protection. We instead regularize in activation space, using a pretrained sparse autoencoder (SAE) as a fixed dictionary of approximately monosemantic features. We formulate continual learning as constrained optimization over feature drift, with a stability constraint on the features to be preserved and a plasticity constraint on the features used by the current task, and we show that the resulting squared-hinge loss is a regularized Lagrangian relaxation of this problem, with EWC as a special case. A replay variant measures drift on 200 stored sequences per previous task, whereas a data-free variant stores only a 197~KB feature mask per task, versus 6.5~GB for EWC. On TRACE (Gemma-2 9B, Llama-3.1 8B) and MedCL (Gemma-2 9B), the replay variant matches or exceeds SAE-based feature distillation at approximately one third of its per-step cost, and the data-free variant attains the highest mean overall performance among methods that store no previous-task data and add no task-specific parameters. Task-relevant units are linearly separable in the SAE basis but not in the neuron basis (ROC-AUC of 0.87--0.88 versus 0.50), and weight-space protection constrains the features of other tasks nearly as strongly as those of the protected task.
\end{abstract}


\section{Introduction}
\label{sec:intro}

\IEEEPARstart{L}{arge} language models (LLMs) increasingly require continual adaptation to new domains, tasks, and interaction formats after deployment. Sequential fine-tuning, however, can overwrite representations needed for earlier capabilities, leading to catastrophic forgetting~\cite{mccloskey1989,ratcliff1990}. Continual learning methods address this through replay, architectural expansion, or regularization. Regularization is particularly appealing for LLMs because it adds no task-specific parameters or inference-time routing and requires little or no stored previous-task data. Most regularizers protect knowledge at the parameter level: Elastic Weight Consolidation (EWC)~\cite{kirkpatrick2017}, Synaptic Intelligence (SI)~\cite{zenke2017}, and Memory Aware Synapses (MAS)~\cite{aljundi2018} assign importance scores to individual weights and penalize later changes to them. This approach is effective when individual parameters are a sufficiently selective unit of protection, but its effectiveness diminishes in LLMs. On TRACE, EWC applied to low-rank adaptation (LoRA)~\cite{hu2022} yields only limited improvement over unprotected sequential fine-tuning~\cite{wang2023trace}. In our experiments, EWC reduces forgetting only at the expense of plasticity: with a penalty weak enough to preserve the learning of new tasks, the model still loses approximately 10\% of each task's score on average by the end of the sequence, whereas a penalty strong enough to nearly eliminate this loss impairs the learning of new tasks far below the level of unprotected fine-tuning. In addition, EWC must store, for each task, a reference copy of the trainable parameters and a per-parameter importance score (together, an \emph{anchor}), $6.5$~GB in our setting.

We argue that part of this limitation is structural rather than a matter of hyperparameter tuning. The superposition hypothesis~\cite{elhage2022} holds that networks encode more features than their dimensionality allows by representing them in overlapping directions. The consequence is polysemanticity: individual neurons and their weights participate in multiple unrelated concepts~\cite{olah2020,bills2023}. Consequently, the diagonal Fisher information used by EWC cannot distinguish whether a weight serves concept~A, concept~B, or both, and protecting it for~A also constrains~B. This mismatch between the unit of importance estimation and the unit of knowledge partly explains why weight-space protection is too coarse for LLMs.

We propose a sparse autoencoder (SAE) as a more selective coordinate system for continual learning. SAEs reconstruct activations through sparse overcomplete codes and yield features that are more often monosemantic, that is, responsive to a single concept, than neurons are~\cite{cunningham2023,bricken2023}. Using a frozen pretrained SAE as a feature dictionary, we estimate which features each task activates and build a graded relevance mask whose high-relevance features form a plasticity set; the features to be preserved form a stability set.

Rather than designing the regularizer heuristically, we derive it from a constrained optimization problem in SAE feature space: minimize the current-task loss subject to (a)~features in the stability set not drifting beyond a stability budget and (b)~features in the plasticity set moving by at least a plasticity floor. The floor excludes the degenerate solution in which all features are preserved and nothing is learned. A regularized Lagrangian relaxation yields two squared-hinge terms, a protect loss and a guide loss, and EWC is recovered as the special case that measures drift in weight space with a zero budget and no plasticity constraint.

The objective admits two variants, which differ mainly in how the stability set is defined and where its drift is measured. The \emph{replay variant} measures drift on a buffer of previous-task activations (200 sequences per task) and constrains the entire dictionary. The \emph{data-free variant} constrains the complement of the plasticity set, measures drift on the current task relative to the base model, and retains only a 197~KB feature mask per task, versus 6.5~GB for an EWC anchor. SAE feature distillation (SAE-FD)~\cite{zhang2026sae} also stores past activations but matches them densely; our replay variant instead imposes a one-sided per-feature constraint at approximately one third of SAE-FD's per-step cost. Neither variant adds per-task parameters or inference-time routing.

We evaluate the method on TRACE (cross-domain) with Gemma-2 9B-it and Llama-3.1-8B-Instruct, and on MedCL (within-domain, biomedical) with Gemma-2 9B-it. On TRACE, the replay variant matches SAE-FD with both models and, with Gemma, attains the strongest retention of any method. On MedCL, with the same settings, it attains the highest mean overall performance and the strongest retention of all methods. In all three settings, the data-free variant has the highest mean overall performance among methods that neither store previous-task data nor add task-specific parameters.\footnote{Its margin over EWC lies within the variation across seeds for Gemma on TRACE; for Llama, EWC's penalty strength is re-selected on the Llama validation split (Appendix~\ref{app:hparams}).} Two analyses support our explanation: task-relevant units are linearly separable in the SAE basis but not in the neuron basis, and protecting weights constrains other tasks' features far more than protecting SAE features does.

Our contributions are:
\begin{itemize}
  \item a feature-space continual learning regularizer built on pretrained SAEs, in a replay variant and a data-free variant of one objective;
  \item a derivation of the squared-hinge loss as the regularized Lagrangian relaxation of an explicit stability--plasticity constrained problem, with EWC as a special case;
  \item evidence that superposition undermines weight-space protection: a separability test (ROC-AUC 0.87--0.88 in the SAE basis versus 0.50 in the neuron basis) and a collateral-constraint test (a collateral-to-intended exposure ratio of 0.91--0.96 for weight protection versus 0.43--0.61 for feature protection);
  \item experiments on two benchmarks and two model families in which the replay variant matches SAE-FD at approximately one third of its per-step cost and the data-free variant has the highest mean overall performance among methods that store no previous-task data and add no task-specific parameters, with a per-task state of 197~KB instead of a 6.5~GB weight anchor.
\end{itemize}

\section{Related Work}

\subsection{Continual Learning and the Stability--Plasticity Tradeoff}

\begin{figure*}[t]
  \centering
  \includegraphics[width=\textwidth]{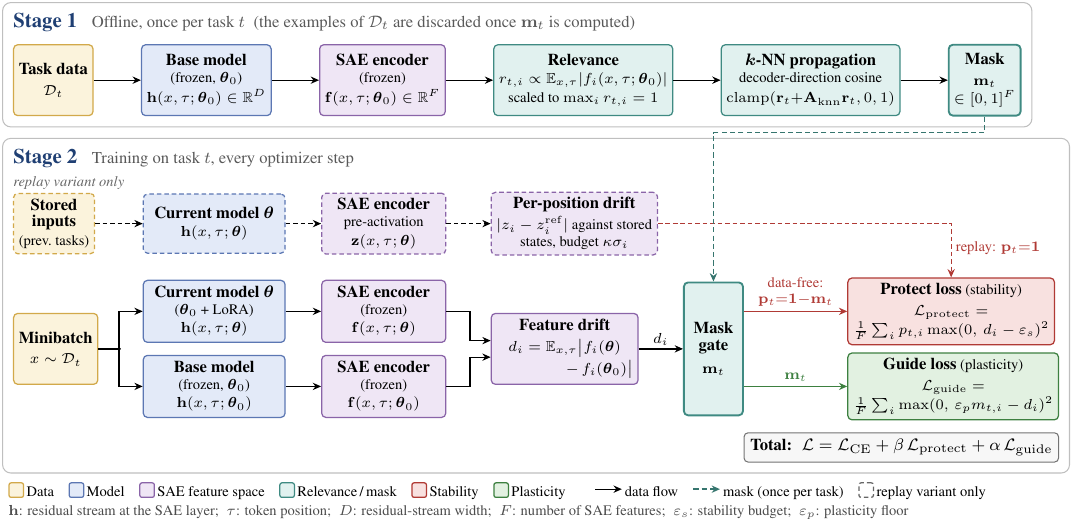}
  \caption{Overview of SAE-guided regularization. \textbf{Stage~1} (offline, per task): current-task data is passed through the frozen base model and frozen SAE to compute a relevance profile, which is expanded by $k$-NN propagation on decoder cosine similarity into a mask $\mathbf{m}_t\in[0,1]^F$; the data used for this step is not retained. \textbf{Stage~2} (per optimizer step): the batch is passed through the current model (with LoRA) and the frozen base model; both activations are encoded by the frozen SAE, and the per-feature drift is gated by the mask into a protect loss on the stability set and a guide loss on the plasticity set. In the data-free variant, the stability score is $\mathbf{1}-\mathbf{m}_t$. In the replay variant (dashed), stored previous-task inputs are also passed through the current model, and their SAE pre-activations are compared position by position with the stored states; this branch replaces the data-free protect loss.}
  \label{fig:workflow}
\end{figure*}

Continual learning requires a model to acquire tasks sequentially without forgetting earlier ones, and hence to remain stable enough to preserve past knowledge yet plastic enough to learn new tasks. Gradient-based training on a task sequence can cause severe forgetting~\cite{mccloskey1989,ratcliff1990}, a problem that is acute for LLMs adapted after deployment to new domains and formats~\cite{wang2023trace,medclbench2026}. Existing methods fall into three families. Replay-based methods store or regenerate previous-task data~\cite{lopezpaz2017,chaudhry2019,buzzega2020}. Architectural methods allocate task-specific parameters, masks, or subspaces~\cite{rusu2016,mallya2018}; for LLMs, O-LoRA~\cite{wang2023olora} trains each task in a LoRA~\cite{hu2022} subspace orthogonal to earlier ones, InfLoRA~\cite{inflora} designs interference-free low-rank subspaces, N-LoRA~\cite{yang2025} reduces parameter collisions through sparsity, and OLieRA~\cite{cao2025} preserves parameter geometry through Lie-group updates. Regularization-based methods keep the architecture fixed and penalize harmful changes~\cite{kirkpatrick2017,zenke2017,aljundi2018}. Strong stability constraints can also cause intransigence, an inability to learn new tasks~\cite{chaudhry2019}, so low forgetting alone does not imply successful continual learning. We treat stability and plasticity as separate requirements and derive the objective from both.

\subsection{Regularization-Based Continual Learning}

Classical regularizers operate in weight space. EWC penalizes weight changes in proportion to the diagonal Fisher information~\cite{kirkpatrick2017}; SI accumulates importance online from each parameter's contribution to loss reduction~\cite{zenke2017}; MAS uses the sensitivity of the output to parameter perturbations~\cite{aljundi2018}; and Benzing~\cite{benzing2022} shows that SI and MAS both approximate the square root of the Fisher information. These methods share the assumption that knowledge can be preserved by constraining important parameters. On LLM continual-learning benchmarks their gains over sequential fine-tuning are modest~\cite{wang2023trace,medclbench2026}, and in our experiments EWC cannot reduce forgetting further without also reducing plasticity (\cref{sec:consequence}). Other methods operate on functions or subspaces. Learning without Forgetting distills an earlier model's outputs~\cite{li2018}; gradient-projection methods restrict updates to directions that leave earlier tasks' representations intact~\cite{gpm}; and ELLA~\cite{ella2026} penalizes alignment of the current LoRA update with the aggregated update of past tasks. Jiang et al.~\cite{jiang_fv} trace forgetting in instruction tuning to shifts in function vectors, compact activation-space representations of a task~\cite{todd2024}, and regularize them. These methods move beyond parameter anchoring, but they protect model outputs, activation subspaces, or task-level vectors; none protects individual concepts while separately requiring the features of the current task to move. Our method changes both the coordinate system, from parameters to SAE features, and the objective, from a one-sided penalty to explicit stability and plasticity constraints.

The work most closely related to ours is SAE-FD~\cite{zhang2026sae}, which also regularizes in SAE feature space through replay-based distillation, matching later SAE features to stored per-token activations with cosine-direction and magnitude losses. Its objective is a dense distillation loss without a plasticity term or a buffer-free mode. Our stability constraint can be measured on the same kind of buffer, as a one-sided per-feature constraint rather than a distillation loss, or on the current task alone, storing nothing from previous tasks.

\subsection{Superposition, Polysemanticity, and Sparse Autoencoders}

Under superposition~\cite{elhage2022}, individual neurons respond to multiple unrelated concepts~\cite{olah2020,bills2023}. SAEs recover more interpretable features from such superposed representations by reconstructing activations through a sparse overcomplete code~\cite{cunningham2023,bricken2023}. SAE features can be identified and manipulated in production-scale models~\cite{templeton2024}, and Gemma Scope provides pretrained SAEs across layers and widths for Gemma~2~\cite{lieberum2024,gemma2}. A recent survey of actionable interpretability organizes interventions as locating an interpretable object and then steering it, with SAE features among the main such objects~\cite{zhang2026locate}; the methods it catalogues intervene almost entirely at inference time. We use the same object but move the intervention into training, as a regularizer on feature drift, so that the deployed model requires no steering mechanism.

\section{Method}
\label{sec:method}

We formulate continual learning as constrained optimization over the internal representation of the model, expressed in the coordinates of a frozen SAE. Because SAE features are approximately monosemantic, the features that a task uses and the features that must be preserved can be distinguished in this basis, whereas they cannot be distinguished in the polysemantic weight or neuron basis (\cref{sec:intro}). The method has two variants, which differ in what is retained from previous tasks and, consequently, in how forgetting is limited. The \emph{data-free variant} retains only a feature mask per task and confines the change made by each task to the features that the task uses. The \emph{replay variant} retains the hidden states of a small number of training sequences from each previous task and keeps the features of these stored inputs unchanged.

The method is developed in four steps. \Cref{sec:setup} represents the hidden states of the model by SAE features and defines the drift of each feature during training. \Cref{sec:scores} assigns to each task a plasticity score and a stability score for every feature, which indicate whether the feature is to be adapted or preserved; the two variants differ in the stability score. \Cref{sec:problem} uses the scores to state a constrained problem, and \cref{sec:formulation} derives the training loss of both variants from this problem and relates it to EWC. \Cref{sec:impl} gives implementation details. \Cref{fig:workflow} summarizes the pipeline, and \cref{alg:training} lists the training procedure.


\subsection{Features and Drift}
\label{sec:setup}

\emph{Setting.} Tasks $t=1,\dots,T$ arrive in sequence, and task $t$ provides a training set $\mathcal{D}_t$ of input sequences in the chat format of the model. Let $\boldsymbol{\theta}_0$ denote the parameters of the frozen pretrained model and $\boldsymbol{\theta}$ the current parameters, which comprise $\boldsymbol{\theta}_0$ and the trainable LoRA weights.

\emph{Hidden states.} In a transformer~\cite{vaswani2017}, each token position is associated with a $D$-dimensional hidden state, to which every layer adds the outputs of its attention and multilayer-perceptron (MLP) blocks through residual connections~\cite{he2016}; this hidden state is therefore termed the \emph{residual stream}~\cite{elhage2021}. For an input $x$, let $\mathbf{h}(x,\tau;\boldsymbol{\theta})\in\mathbb{R}^D$ denote its value at token position $\tau$ after a chosen layer. Positions occupied by chat-template markers, padding, and the end-of-sequence token are excluded from all averages below; we denote by $\mathcal{C}(x)$ the remaining positions of $x$, which hold the text of the example.

\emph{SAE features.} An SAE trained on the hidden states of $\boldsymbol{\theta}_0$ at this layer encodes $\mathbf{h}(x,\tau;\boldsymbol{\theta})$ into a sparse feature vector
\begin{equation}
  \mathbf{f}(x,\tau;\boldsymbol{\theta})=\sigma\bigl(\mathbf{z}(x,\tau;\boldsymbol{\theta})\bigr)\in\mathbb{R}^F,
  \label{eq:sae_encode}
\end{equation}
where the encoder pre-activation $\mathbf{z}$ is an affine function of $\mathbf{h}(x,\tau;\boldsymbol{\theta})$ and $\sigma$ is the JumpReLU nonlinearity of the SAEs used here~\cite{rajamanoharan2024}, which sets each coordinate below a learned threshold to zero. We call each coordinate $f_i$ a \emph{feature}. A linear decoder reconstructs the activation, up to a constant bias, as $\sum_{i=1}^{F}f_i\mathbf{w}_i$, so the decoder direction $\mathbf{w}_i\in\mathbb{R}^D$ is the vector that feature~$i$ adds to the reconstruction per unit of activation. These $F$ directions form the dictionary; its size $F$, the SAE width, exceeds $D$, so the dictionary is overcomplete. The SAE is frozen throughout training and serves only as a fixed coordinate system for measuring changes to the model. When several layers are regularized, each has its own SAE and all quantities below are per layer.

\emph{Drift.} During training, the value of each feature changes. We measure this change relative to a reference value $f_i^{\mathrm{ref}}(x,\tau)$ (\cref{fig:workflow}, Stage~2), which, unless stated otherwise, is the value of the feature under the base model, $f_i^{\mathrm{ref}}(x,\tau)=f_i(x,\tau;\boldsymbol{\theta}_0)$. For a set $\mathcal{S}$ of inputs, we write $(x,\tau)\in\mathcal{S}$ for $x\in\mathcal{S}$ and $\tau\in\mathcal{C}(x)$, and $\mathbb{E}_{(x,\tau)\in\mathcal{S}}$ for the average over all such pairs. The \emph{mean drift} of feature $i$ on $\mathcal{S}$ is the average magnitude of its change,
\begin{equation}
  d_i(\boldsymbol{\theta};\mathcal{S})=\mathbb{E}_{(x,\tau)\in\mathcal{S}}\bigl|f_i(x,\tau;\boldsymbol{\theta})-f_i^{\mathrm{ref}}(x,\tau)\bigr|,
  \label{eq:mean_drift}
\end{equation}
where the absolute value counts changes of either sign.

\subsection{Plasticity and Stability Scores}
\label{sec:scores}

\begin{figure}[t]
  \centering
  \includegraphics[width=\columnwidth]{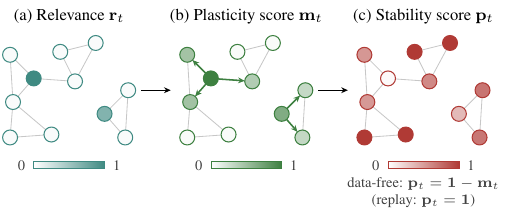}
  \caption{Illustration of the plasticity and stability scores. Nodes are features placed by the similarity of their decoder directions, gray edges form the $k$-nearest-neighbor graph ($k{=}2$ for illustration), and shading gives the score of each feature. (a)~Two features are active on task~$t$. (b)~The propagation \eqref{eq:adjacency} gives their neighbors intermediate plasticity scores (arrows). (c)~The data-free stability score \eqref{eq:replayfree} fully preserves the features unrelated to the task; in the replay variant, $p_{t,i}=1$ for every feature.}
  \label{fig:scores}
\end{figure}

Each task $t$ is described by two score vectors (\cref{fig:scores}): a plasticity score $\mathbf{m}_t\in[0,1]^F$ (the mask of \cref{fig:workflow}), whose entry $m_{t,i}$ measures how strongly task $t$ uses feature $i$, and a stability score $\mathbf{p}_t\in[0,1]^F$, whose entry $p_{t,i}$ measures how strongly feature $i$ is to be preserved during task $t$.

\emph{Plasticity scores.} The plasticity score is computed from the base model $\boldsymbol{\theta}_0$ in two steps (\cref{fig:workflow}, Stage~1). First, the use of each feature by the task is estimated by its mean absolute activation on $\mathcal{D}_t$,
\begin{equation}
  r_{t,i}\propto\mathbb{E}_{(x,\tau)\in\mathcal{D}_t}\bigl|f_i(x,\tau;\boldsymbol{\theta}_0)\bigr|,
  \label{eq:relevance}
\end{equation}
scaled so that the largest entry of the relevance vector $\mathbf{r}_t=(r_{t,1},\dots,r_{t,F})$ is~1 (\cref{fig:scores}\textcolor{figref}{a}). Second, the relevance is extended to related features, because features that the training examples activate only weakly may still be used by the task. Since the decoder direction determines what a feature adds to the residual stream, features with similar decoder directions have similar effects on later layers and are likely to represent related concepts. We call the $k$ features whose decoder directions have the largest cosine similarity to $\mathbf{w}_i$ the \emph{neighbors} of feature $i$, denoted $\mathcal{N}(i)$, and add to the relevance of each feature the relevance of its neighbors, weighted by their similarity:
\begin{equation}
  \mathbf{m}_t=\mathrm{clamp}\bigl(\mathbf{r}_t+\mathbf{A}_{\mathrm{knn}}\mathbf{r}_t,\;0,\;1\bigr),
  \label{eq:adjacency}
\end{equation}
where $\mathbf{A}_{\mathrm{knn}}\in\mathbb{R}^{F\times F}$ has entries $\max\bigl(0,\cos(\mathbf{w}_i,\mathbf{w}_j)\bigr)$ for $j\in\mathcal{N}(i)$ and zero otherwise, so that the $i$th entry of $\mathbf{A}_{\mathrm{knn}}\mathbf{r}_t$ is the similarity-weighted relevance of the neighbors of feature $i$, and $\mathrm{clamp}(\cdot,0,1)$ clips each entry to $[0,1]$. The resulting scores are close to~1 for the features that the task uses, intermediate for their neighbors, and close to~0 elsewhere (\cref{fig:scores}\textcolor{figref}{b}).

\begin{figure}[t]
  \centering
  \includegraphics[width=\columnwidth]{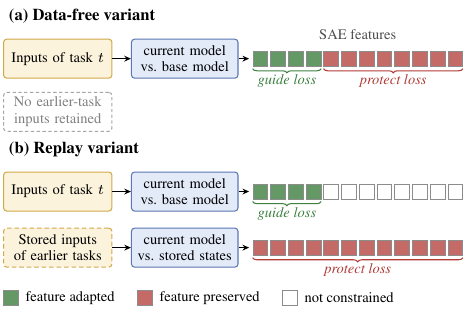}
  \caption{Difference between the two variants during training on task~$t$ (schematic with binary scores; each square is an SAE feature). Each row corresponds to one set of inputs, which is passed through the current model and a reference model; the drift of each feature is penalized by the loss indicated below it. (a)~The data-free variant has only the inputs of task~$t$, on which the features that the task uses are adapted and all other features are preserved. (b)~The replay variant also has stored inputs of earlier tasks: the inputs of task~$t$ serve only the guide loss, and the stored inputs serve the protect loss, which preserves every feature on them.}
  \label{fig:variants}
\end{figure}

\emph{Stability scores.} In addition to the stability score, the stability constraint requires two choices: the set $\mathcal{S}_t$ of inputs on which the drift of a feature is measured, and the reference against which it is measured. All three depend on whether data from previous tasks are retained, and they define the two variants of the method (\cref{fig:variants}).

In the \emph{data-free variant}, no data from previous tasks are retained, so the features that earlier tasks use cannot be identified. The variant instead restricts the change induced by the current task to the features that this task uses: every other feature is preserved, and its drift is measured on the current task relative to the base model,
\begin{equation}
  \mathbf{p}_t=\mathbf{1}-\mathbf{m}_t,\quad \mathcal{S}_t=\mathcal{D}_t,\quad f_i^{\mathrm{ref}}(x,\tau)=f_i(x,\tau;\boldsymbol{\theta}_0),
  \label{eq:replayfree}
\end{equation}
where $\mathbf{1}\in\mathbb{R}^F$ is the all-ones vector (\cref{fig:scores}\textcolor{figref}{c}). The training examples are discarded once the scores are computed.

In the \emph{replay variant}, after each task $s$ is learned, the hidden states of a subset of its training sequences are stored, computed with the parameters $\boldsymbol{\theta}_s^\ast$ reached at the end of that task. Any change of the features on these inputs is potential forgetting, so every feature is preserved on them: $\mathbf{p}_t=\mathbf{1}$, $\mathcal{S}_t$ is the set of inputs stored after tasks $1,\dots,t-1$, and the reference for an input stored after task $s$ is the value of the feature when it was stored, $f_i^{\mathrm{ref}}(x,\tau)=f_i(x,\tau;\boldsymbol{\theta}_s^\ast)$.

\subsection{Constrained Formulation}
\label{sec:problem}

To derive the training loss, we first treat the scores of \cref{sec:scores} as binary: the features with $p_{t,i}=1$ form the \emph{stability set}, which is to be preserved, and those with $m_{t,i}=1$ form the \emph{plasticity set}, which the task uses. At task $t$, we seek parameters that minimize the task loss while preserving the features in the stability set and adapting those in the plasticity set. Expressing both requirements through the mean drift \eqref{eq:mean_drift} yields the constrained problem
\begin{align}
  \min_{\boldsymbol{\theta}}\quad & \mathcal{L}_{\mathrm{CE}}(\boldsymbol{\theta};\mathcal{D}_t) \label{eq:obj}\\
  \text{s.t.}\quad & d_i(\boldsymbol{\theta};\mathcal{S}_t)\le\varepsilon_s,\quad \forall\, i:\,p_{t,i}=1, \label{eq:stability}\\
  & d_i(\boldsymbol{\theta};\mathcal{D}_t)\ge\varepsilon_p m_{t,i},\quad \forall\, i. \label{eq:plasticity}
\end{align}
Here $\mathcal{L}_{\mathrm{CE}}(\boldsymbol{\theta};\mathcal{D}_t)$ is the next-token cross-entropy on $\mathcal{D}_t$, written $\mathcal{L}_{\mathrm{CE}}(\boldsymbol{\theta})$ below. The stability budget $\varepsilon_s$ is the largest mean drift allowed for a feature in the stability set (a ceiling on drift), and the plasticity floor $\varepsilon_p$ is the smallest mean drift required of a feature in the plasticity set; for $m_{t,i}=0$, \eqref{eq:plasticity} reduces to $d_i\ge0$ and imposes nothing. Both are calibrated on the drift observed when the first task is fine-tuned without regularization (Appendix~\ref{app:hparams}). In the data-free variant, $\varepsilon_s$ also leaves room for features that earlier tasks moved outside the current plasticity set. The plasticity constraint \eqref{eq:plasticity} concerns the task being learned and is therefore measured on $\mathcal{D}_t$.

\emph{Rationale for the constraints.} The plasticity constraint does not prescribe the direction of change, which is determined by the task loss; it only excludes the degenerate solution in which no feature changes. Both constraints are imposed per feature because an aggregate constraint permits compensation across features: an average budget can be satisfied while a few features drift substantially, and an average floor can be met by changing only a few features.

\subsection{Training Loss}
\label{sec:formulation}

We derive the training loss from \eqref{eq:obj}--\eqref{eq:plasticity} in three steps: we form the Lagrangian, eliminate its multipliers in closed form, and extend the result to continuous scores. We then adapt the stability term to the replay variant.

\emph{Lagrangian.} With multipliers $\mu_i\ge0$ for the stability constraints and $\nu_i\ge0$ for the plasticity constraints, collected in the vectors $\boldsymbol{\mu}$ and $\boldsymbol{\nu}$, the Lagrangian of \eqref{eq:obj}--\eqref{eq:plasticity} is
\begin{equation}
\begin{split}
  \mathcal{J}(\boldsymbol{\theta},\boldsymbol{\mu},\boldsymbol{\nu})={}&\mathcal{L}_{\mathrm{CE}}(\boldsymbol{\theta})+\sum_{i:\,p_{t,i}=1}\mu_i\bigl(d_i(\boldsymbol{\theta};\mathcal{S}_t)-\varepsilon_s\bigr)\\
  &+\sum_{i=1}^{F}\nu_i\bigl(\varepsilon_p m_{t,i}-d_i(\boldsymbol{\theta};\mathcal{D}_t)\bigr).
\end{split}
\label{eq:lagrangian}
\end{equation}
The treatment of the multipliers is critical. Fixing each at a constant reduces \eqref{eq:lagrangian} to a linear penalty on drift whose gradient does not depend on whether the constraint is satisfied: features within their budget are still driven toward zero drift, and $\varepsilon_s$ has no effect on the solution. Optimizing the multipliers jointly with $\boldsymbol{\theta}$ enforces the constraints but introduces up to $2F$ additional variables and a second optimization problem that is often unstable. We therefore subtract a quadratic cost on the multipliers from \eqref{eq:lagrangian} and maximize over them, so that each multiplier becomes a function of the current violation and vanishes when its constraint is satisfied.

\emph{Closed-form multipliers.} Because both the Lagrangian and the cost separate over features, it suffices to solve the maximization for a single constraint. Consider a stability constraint, let $d_i$ denote $d_i(\boldsymbol{\theta};\mathcal{S}_t)$, and assign its multiplier the cost $a\mu_i^2$ with $a>0$, the simplest smooth cost for which the maximizing multiplier is finite. Then
\begin{equation}
  \max_{\mu_i\ge0}\bigl[\mu_i(d_i-\varepsilon_s)-a\mu_i^2\bigr]=\frac{1}{4a}\max(0,\,d_i-\varepsilon_s)^2,
  \label{eq:scalar}
\end{equation}
attained at $\mu_i=\max(0,\,d_i-\varepsilon_s)/(2a)$, which is zero when the constraint is satisfied and grows linearly with the violation otherwise. We set $a=F/(4\beta)$ for every stability multiplier and $a=F/(4\alpha)$ for every plasticity multiplier, so that $\beta$ and $\alpha$ become the weights of penalties averaged over the $F$ dictionary features. The regularized Lagrangian is then
\begin{equation}
  \mathcal{L}(\boldsymbol{\theta})=\max_{\boldsymbol{\mu},\boldsymbol{\nu}\ge0}\Bigl[\mathcal{J}(\boldsymbol{\theta},\boldsymbol{\mu},\boldsymbol{\nu})-\frac{F}{4\beta}\|\boldsymbol{\mu}\|^{2}-\frac{F}{4\alpha}\|\boldsymbol{\nu}\|^{2}\Bigr].
  \label{eq:regdual}
\end{equation}
Substituting \eqref{eq:lagrangian} into \eqref{eq:regdual}, the term $\mathcal{L}_{\mathrm{CE}}$ does not depend on the multipliers, and the remaining terms separate into one-multiplier problems of the form \eqref{eq:scalar}, with $\varepsilon_p m_{t,i}-d_i(\boldsymbol{\theta};\mathcal{D}_t)$ replacing $d_i-\varepsilon_s$ for the plasticity terms. Summing their maxima, and writing the sum over the stability set as a sum over all features weighted by the binary $p_{t,i}$, yields
\begin{equation}
  \mathcal{L}(\boldsymbol{\theta})=\mathcal{L}_{\mathrm{CE}}(\boldsymbol{\theta})+\beta\,\mathcal{L}_{\mathrm{protect}}+\alpha\,\mathcal{L}_{\mathrm{guide}},
  \label{eq:hinge}
\end{equation}
with the dictionary-averaged penalties
\begin{align}
  \mathcal{L}_{\mathrm{protect}}&=\frac{1}{F}\sum_{i=1}^{F} p_{t,i}\max\bigl(0,\,d_i(\boldsymbol{\theta};\mathcal{S}_t)-\varepsilon_s\bigr)^2, \label{eq:lprotect}\\
  \mathcal{L}_{\mathrm{guide}}&=\frac{1}{F}\sum_{i=1}^{F} \max\bigl(0,\,\varepsilon_p m_{t,i}-d_i(\boldsymbol{\theta};\mathcal{D}_t)\bigr)^2. \label{eq:lguide}
\end{align}
The loss \eqref{eq:hinge} is therefore not a heuristic penalty: it is the Lagrangian \eqref{eq:lagrangian} with its multipliers maximized in closed form, that is, a regularized Lagrangian relaxation of \eqref{eq:obj}--\eqref{eq:plasticity}.\footnote{Equivalently, $\mathcal{L}$ is the quadratic-penalty term of the augmented Lagrangian at a zero multiplier estimate~\cite{nocedal2006}. Replacing the quadratic cost in \eqref{eq:regdual} by a box constraint on the multipliers gives the linear hinge, the classical exact penalty; we use the squared form for its continuous gradient.} The weights have a direct interpretation. As $\beta\to\infty$, the cost on the stability multipliers vanishes and the hard constraint \eqref{eq:stability} is recovered, whereas $\beta\to0$ removes the constraint; $\alpha$ acts in the same way on \eqref{eq:plasticity}. Because both penalties are dictionary averages, $\alpha$ and $\beta$ remain comparable across SAE widths.

\begin{figure}[t]
  \centering
  \includegraphics[width=\columnwidth]{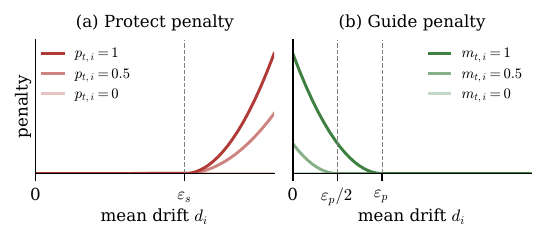}
  \caption{Illustration of the per-feature penalties as functions of the mean drift $d_i$; the thresholds are not to scale. (a)~The protect penalty \eqref{eq:lprotect} is zero up to the stability budget $\varepsilon_s$ and quadratic beyond it; the stability score $p_{t,i}$ scales its height. (b)~The guide penalty \eqref{eq:lguide} is zero above the floor $\varepsilon_p m_{t,i}$ and quadratic below it; the plasticity score shifts the floor rather than scaling the penalty. Curves for a score of~0 lie on the horizontal axis.}
  \label{fig:penalty}
\end{figure}

\emph{Continuous scores.} The derivation assumes binary scores, whereas the scores of \cref{sec:scores} are continuous: the relevance \eqref{eq:relevance} is a scaled mean activation, and its propagation \eqref{eq:adjacency} takes values throughout $[0,1]$. Rather than thresholding them, which would introduce a cutoff hyperparameter, we use \eqref{eq:hinge}--\eqref{eq:lguide} unchanged with the continuous scores, and each score retains a constrained interpretation (\cref{fig:penalty}). In $\mathcal{L}_{\mathrm{protect}}$, $p_{t,i}$ weights the penalty of feature $i$, which corresponds to the multiplier cost $a=F/(4\beta p_{t,i})$ in \eqref{eq:scalar}; the budget $\varepsilon_s$ is unchanged, and a feature with $p_{t,i}=0$ is unconstrained. In $\mathcal{L}_{\mathrm{guide}}$, $m_{t,i}$ scales the floor, so that the required drift is proportional to the relevance of the feature; weighting the guide penalty by $m_{t,i}$ instead would demand the full drift $\varepsilon_p$ from every weakly relevant feature. In \cref{fig:workflow}, both scores are applied by the mask gate.

\emph{Replay variant.} In the replay variant, the stored hidden states provide the reference value of every feature at every stored position, so the stability constraint can be imposed at each stored position rather than on the mean drift as in \eqref{eq:stability}. Two further changes adapt it to the stored data. First, the constraint is placed on the pre-activation $z_i$ rather than on the feature $f_i$: for a feature that is inactive on a stored input, $f_i$ is zero and does not reflect how close the feature is to activating, whereas $z_i$ does, so the same constraint covers the drift of active features and the spurious activation of inactive ones. Second, because activation scales differ widely across features, each feature receives its own budget $\kappa\sigma_i$, where $\sigma_i$ is the standard deviation of its stored pre-activations and $\kappa>0$. With $z_i^{\mathrm{ref}}(x,\tau)=z_i(x,\tau;\boldsymbol{\theta}_s^\ast)$ the stored pre-activation of an input $x$ stored after task $s$, the stability constraints of the replay variant are
\begin{equation}
  \bigl|z_i(x,\tau;\boldsymbol{\theta})-z_i^{\mathrm{ref}}(x,\tau)\bigr|\le\kappa\sigma_i,\quad \forall\, i,\ (x,\tau)\in\mathcal{S}_t.
  \label{eq:replaycon}
\end{equation}
These constraints are relaxed in the same way as \eqref{eq:stability}. Let $N_t$ be the number of stored positions in $\mathcal{S}_t$. Assigning the multiplier of each constraint the cost $a=FN_t\sigma_i^2/(4\beta)$ makes $\beta$ the weight of an average over features and positions and measures each violation in units of $\sigma_i$; by \eqref{eq:scalar}, the protect loss \eqref{eq:lprotect} is then replaced by
\begin{equation}
\begin{split}
  \mathcal{L}_{\mathrm{protect}}={}&\frac{1}{F}\sum_{i=1}^{F}\frac{1}{\sigma_i^2}\,\mathbb{E}_{(x,\tau)\in\mathcal{S}_t}\\
  &\max\bigl(0,\,|z_i(x,\tau;\boldsymbol{\theta})-z_i^{\mathrm{ref}}(x,\tau)|-\kappa\sigma_i\bigr)^2.
\end{split}
\label{eq:lreplay}
\end{equation}
In \cref{fig:workflow}, the replay variant adds a Stage-2 branch for the stored inputs, which feeds only the protect loss. Both variants thus minimize a loss of the form \eqref{eq:hinge} with the same guide loss \eqref{eq:lguide} and differ only in the stability term.

\emph{Relation to EWC.} EWC is the special case of \eqref{eq:hinge} in which the coordinates are the weights, the reference is the previous checkpoint, the stability score is the diagonal Fisher information, the budget is zero, and there is no plasticity constraint (Appendix~\ref{app:ewc}). Our method departs from EWC in the coordinates, the budget, and the plasticity constraint, which are evaluated by the separability and collateral analyses (\cref{sec:separability,sec:collateral}) and the term ablation (\cref{sec:factorial}).

\subsection{Implementation Details}
\label{sec:impl}

\begin{algorithm}[t]
\caption{Training on task $t$}
\label{alg:training}
\begin{algorithmic}[1]
\Require training set $\mathcal{D}_t$; frozen base model $\boldsymbol{\theta}_0$ and SAE; current parameters $\boldsymbol{\theta}$; for replay, the buffer of previous tasks
\Statex \textit{Stage~1: scores (\cref{sec:scores})}
\State $\mathbf{r}_t\gets$ relevance on $\mathcal{D}_t$ \Comment{\eqref{eq:relevance}}
\State $\mathbf{m}_t\gets\mathrm{clamp}(\mathbf{r}_t+\mathbf{A}_{\mathrm{knn}}\mathbf{r}_t,0,1)$ \Comment{\eqref{eq:adjacency}}
\State \textbf{if} data-free \textbf{then} $\mathbf{p}_t\gets\mathbf{1}-\mathbf{m}_t$ \Comment{\eqref{eq:replayfree}}
\Statex \textit{Stage~2: training (\cref{sec:formulation})}
\For{each minibatch $\mathcal{X}\subset\mathcal{D}_t$}
  \State encode $\mathbf{h}(\cdot;\boldsymbol{\theta})$ and $\mathbf{h}(\cdot;\boldsymbol{\theta}_0)$ on $\mathcal{X}$
  \State $\mathcal{L}_{\mathrm{guide}}\gets$ \eqref{eq:lguide} with drift on $\mathcal{X}$
  \If{data-free}
    \State $\mathcal{L}_{\mathrm{protect}}\gets$ \eqref{eq:lprotect} with $\mathcal{S}_t=\mathcal{X}$
  \Else
    \State sample four stored sequences
    \State $\mathcal{L}_{\mathrm{protect}}\gets$ \eqref{eq:lreplay}; cap $\beta$
  \EndIf
  \State update LoRA weights on $\mathcal{L}_{\mathrm{CE}}+\beta\mathcal{L}_{\mathrm{protect}}+\alpha\mathcal{L}_{\mathrm{guide}}$
\EndFor
\State \textbf{if} replay \textbf{then} store hidden states of 200 sequences of $\mathcal{D}_t$
\end{algorithmic}
\end{algorithm}

The $k$-nearest-neighbor graph in \eqref{eq:adjacency} uses $k{=}10$. The expectations in \eqref{eq:mean_drift} and \eqref{eq:lreplay} are replaced by minibatch means, computed at each optimizer step from forward passes of the current and base models through the SAE encoder. When several layers are regularized, the two losses are computed per layer, with layer-specific budgets, and summed.

For the replay variant, after each task we store the token IDs and the hidden states of the deepest regularized layer at every position for $200$ of its training sequences, computed with the weights obtained at the end of that task and kept in 16-bit precision (approximately $1.4$~GB per task at layer~31 of Gemma-2 9B). SAE features are not stored; the hidden states are encoded on demand, so the buffer is independent of the regularizer's hyperparameters. At each optimizer step, four stored sequences are sampled uniformly from all previous tasks. The replay protect loss is applied at that layer only, and $\beta$ is capped at each step so that the protect term contributes at most half of the task loss. The buffer size, the number of sampled sequences, and $\kappa{=}1$ were fixed a priori rather than tuned.
\section{Experiments}
\label{sec:experiments}

We evaluate the method on two continual-learning benchmarks and two model families. The experiments address four questions: (i)~whether regularization in the SAE feature basis improves on weight-space regularization when no previous-task data are stored, which concerns the data-free variant; (ii)~how the replay variant compares with methods that store previous-task activations or examples; (iii)~what the method costs in computation, memory, and storage; and (iv)~how sensitive the method is to its hyperparameters and which loss terms produce its effect. After the setup (\cref{sec:setup_exp}), \cref{sec:main_results} addresses questions~(i) and~(ii), \cref{sec:cost} question~(iii), and \cref{sec:sweep,sec:factorial} question~(iv). \Cref{sec:framework} then examines why regularization in the feature basis is more selective.

\subsection{Setup}
\label{sec:setup_exp}

\emph{Benchmarks.} TRACE-5000~\cite{wang2023trace} is a fixed sequence of eight cross-domain tasks with 5,000 training examples each: C-STANCE (Chinese stance detection), FOMC (financial sentiment), MeetingBank (summarization), Py150 (code completion), ScienceQA (science question answering), NumGLUE-cm and NumGLUE-ds (arithmetic reasoning), and 20Minuten (German text simplification). We follow its protocol, with per-task epochs $[5,3,7,5,3,5,5,7]$ and learning rate $10^{-4}$. MedCL~\cite{medclbench2026} is a sequence of ten biomedical tasks (PubMedQA, SciFact, PubHealth, GAD, ChemProt, DDI, PubMed RCT, LitCovid, BioASQ, and DrugLib) that share a domain but differ in format and label vocabulary; it complements TRACE by testing interference within a single domain. We train for three epochs per task at learning rate $5\times10^{-5}$. Because the label vocabularies of MedCL are short, the model is evaluated by ranking the candidate labels by their log-probability, which avoids the sensitivity of open-ended decoding to output format.

\emph{Models.} The primary model is Gemma-2 9B-it~\cite{gemma2} with LoRA~\cite{hu2022} on all linear projections ($r{=}256$, $\alpha_{\mathrm{LoRA}}{=}512$, dropout 0.05). The regularizer acts on the residual stream at layers 9, 20, and 31, each with a frozen Gemma Scope SAE~\cite{lieberum2024} of width $F{=}16{,}384$. To test transfer to another model family, we use Llama-3.1-8B-Instruct~\cite{llama3} on TRACE with the same LoRA configuration. The public SAEs for this model reconstructed only approximately 57\% of the activation variance on our data; we therefore trained a JumpReLU SAE~\cite{rajamanoharan2024} on the activations of the instruct model at layer~20 (approximately 90\% explained variance).

\emph{Baselines.} The baselines are grouped by the information they retain from previous tasks, so that each variant of our method can be compared with methods of matching storage. (1)~\emph{Methods that store no previous-task data}: the weight-space regularizers EWC~\cite{kirkpatrick2017}, SI~\cite{zenke2017}, and MAS~\cite{aljundi2018}, which keep per-parameter state, and ELLA~\cite{ella2026}, which penalizes alignment with the aggregated past LoRA update. The data-free variant belongs to this group, and ``regularization-based methods'' refers to it below. (2)~\emph{Activation-buffer methods}, which store previous-task hidden states: SAE-FD~\cite{zhang2026sae} and the replay variant. (3)~\emph{Example replay}, which mixes 2\% or 10\% of the training set of each previous task into later training. (4)~\emph{Architectural isolation}: O-LoRA~\cite{wang2023olora} allocates an orthogonal rank-32 LoRA subspace per task and thus adds parameters with every task; it serves as a reference outside the regularization setting. Unprotected sequential fine-tuning serves as the lower reference.

\emph{Training and hyperparameters.} All methods use 8-bit AdamW~\cite{loshchilov2019,dettmers2022} with an effective batch size of 128 and the learning rates given above; all except O-LoRA use the LoRA configuration given above. Hyperparameters are selected on the TRACE validation split with Gemma, which is disjoint from the test split on which all results are reported, and carried over unchanged to MedCL. On Llama, both our method and EWC are re-tuned on the same validation split (Appendix~\ref{app:hparams}). For our method, a reference configuration is first set from the drift observed when the first task is fine-tuned without regularization: $\varepsilon_s$ and $\varepsilon_p$ at the 90th and 25th percentiles of this drift, and $\alpha$ and $\beta$ such that each loss term equals 5\% of the cross-entropy at this drift level ($\alpha{=}480$ and $\beta{=}6.5$ for Gemma on TRACE). Only $\alpha$ is then selected on validation. Llama uses a different SAE, so its reference configuration is recomputed in the same way and its ceiling weight $\beta$ is selected on validation. Among the baselines, only the weight $\lambda$ of EWC is selected on validation, on both models. MAS and ELLA use the same $\lambda$, with their importance scores rescaled to the mean magnitude of EWC's Fisher information so that the weight acts on a comparable scale; SI uses its standard settings, and O-LoRA and SAE-FD use the settings of their original papers. The settings of every method on each benchmark and model, the selection protocol, and the validation scores are given in Appendix~\ref{app:hparams}.

\emph{Metrics.} Let $\mathrm{diag}_j$ denote the score on task $j$ immediately after training on it and $\mathrm{fin}_j$ the score after training on the whole sequence. Overall performance (OP) is the mean of $\mathrm{fin}_j$; backward transfer (BWT) is the mean of $\mathrm{fin}_j-\mathrm{diag}_j$; average retained accuracy (ARA), our measure of retention, is the mean of $\min(\mathrm{fin}_j/\mathrm{diag}_j,1)$; and plasticity is the mean of $\mathrm{diag}_j$, all over the $T$ tasks, including the last. The per-task score is accuracy for the classification and arithmetic tasks of TRACE and for all MedCL tasks, ROUGE-L F1 for MeetingBank and 20Minuten, and normalized character-level edit similarity for Py150. Main results are means over three training seeds, with OP reported as mean $\pm$ standard deviation (SD). We report the spread over seeds rather than bootstrap intervals over evaluation examples because training variability dominates the uncertainty, by approximately a factor of three for the least stable methods. The sensitivity and ablation studies use a single seed.

\subsection{Main Results}
\label{sec:main_results}

\begin{table}[t]
\centering
\caption{TRACE-5000 results (Gemma-2 9B-it, 8 tasks). Three-seed means; OP as mean $\pm$ sample SD. Rows are grouped by what is retained from previous tasks and ranked by OP within each group. $^\dagger$Stores no previous-task data or activations.}
\label{tab:trace}
\footnotesize
\setlength{\tabcolsep}{3.2pt}
\begin{tabular}{@{}lcccc@{}}
\toprule
Method & OP $\uparrow$ & ARA $\uparrow$ & BWT $\uparrow$ & Plast. $\uparrow$ \\
\midrule
\textbf{SAE data-free$^\dagger$ (ours)} & \textbf{0.539 $\pm$ 0.015} & \textbf{0.932} & \textbf{$-$0.041} & \textbf{0.580} \\
EWC$^\dagger$ & 0.524 $\pm$ 0.016 & 0.896 & $-$0.060 & 0.584 \\
SI$^\dagger$ & 0.509 $\pm$ 0.020 & 0.891 & $-$0.063 & 0.572 \\
MAS$^\dagger$ & 0.495 $\pm$ 0.038 & 0.863 & $-$0.087 & 0.582 \\
ELLA$^\dagger$ & 0.421 $\pm$ 0.052 & 0.728 & $-$0.154 & 0.575 \\
\midrule
SAE-FD (act.\ buffer) & 0.565 $\pm$ 0.025 & 0.956 & $-$0.025 & 0.591 \\
\textbf{SAE + replay (ours)} & \textbf{0.559 $\pm$ 0.009} & \textbf{0.970} & \textbf{$-$0.012} & \textbf{0.570} \\
\midrule
Example replay 10\% & 0.542 $\pm$ 0.014 & 0.932 & $-$0.036 & 0.576 \\
Example replay 2\% & 0.534 $\pm$ 0.009 & 0.920 & $-$0.044 & 0.577 \\
\midrule
O-LoRA$^\dagger$ (architectural) & 0.627 $\pm$ 0.008 & 0.965 & $-$0.021 & 0.648 \\
\midrule
Unprotected & 0.379 $\pm$ 0.032 & 0.693 & $-$0.185 & 0.564 \\
\bottomrule
\end{tabular}
\end{table}

\begin{table}[t]
\centering
\caption{MedCL results (Gemma-2 9B-it, 10 biomedical tasks, test split, log-probability evaluation). Three-seed means; OP as mean $\pm$ SD. Rows are grouped and ranked as in \cref{tab:trace}. $^\dagger$Stores no previous-task data or activations.}
\label{tab:medcl}
\footnotesize
\setlength{\tabcolsep}{3.2pt}
\begin{tabular}{@{}lcccc@{}}
\toprule
Method & OP $\uparrow$ & ARA $\uparrow$ & BWT $\uparrow$ & Plast. $\uparrow$ \\
\midrule
\textbf{SAE data-free$^{\dagger}$ (ours)} & \textbf{0.501 $\pm$ 0.023} & \textbf{0.805} & \textbf{$-$0.134} & \textbf{0.635} \\
EWC$^\dagger$ & 0.477 $\pm$ 0.021 & 0.747 & $-$0.176 & 0.653 \\
MAS$^\dagger$ & 0.466 $\pm$ 0.017 & 0.751 & $-$0.174 & 0.640 \\
\midrule
\textbf{SAE + replay (ours)} & \textbf{0.620 $\pm$ 0.009} & \textbf{0.926} & \textbf{$-$0.042} & \textbf{0.662} \\
SAE-FD (act.\ buffer) & 0.519 $\pm$ 0.015 & 0.832 & $-$0.105 & 0.623 \\
\midrule
O-LoRA$^\dagger$ (architectural) & 0.509 $\pm$ 0.011 & 0.863 & $-$0.067 & 0.576 \\
\midrule
Unprotected & 0.482 $\pm$ 0.010 & 0.745 & $-$0.185 & 0.667 \\
\bottomrule
\end{tabular}
\end{table}

\begin{table}[t]
\centering
\caption{TRACE-5000 results on a second model family (Llama-3.1-8B-Instruct, 8 tasks). Three-seed means; OP as mean $\pm$ SD. Rows are grouped and ranked as in \cref{tab:trace}. $^\dagger$Stores no previous-task data or activations.}
\label{tab:trace_llama}
\footnotesize
\setlength{\tabcolsep}{3.2pt}
\begin{tabular}{@{}lcccc@{}}
\toprule
Method & OP $\uparrow$ & ARA $\uparrow$ & BWT $\uparrow$ & Plast. $\uparrow$ \\
\midrule
\textbf{SAE data-free$^\dagger$ (ours)} & \textbf{0.575 $\pm$ 0.006} & \textbf{0.930} & \textbf{$-$0.045} & \textbf{0.620} \\
EWC$^\dagger$ & 0.564 $\pm$ 0.002 & 0.973 & $-$0.014 & 0.578 \\
\midrule
\textbf{SAE + replay (ours)} & \textbf{0.601 $\pm$ 0.004} & \textbf{0.979} & \textbf{$-$0.011} & \textbf{0.613} \\
SAE-FD (act.\ buffer) & 0.597 $\pm$ 0.008 & 0.981 & $-$0.009 & 0.606 \\
\midrule
O-LoRA$^\dagger$ (architectural) & 0.602 $\pm$ 0.005 & 0.956 & $-$0.027 & 0.628 \\
\midrule
Unprotected & 0.556 $\pm$ 0.015 & 0.918 & $-$0.051 & 0.607 \\
\bottomrule
\end{tabular}
\end{table}

Two findings hold on both benchmarks and both models (\cref{tab:trace,tab:medcl,tab:trace_llama}). First, among the methods that neither store previous-task data nor add task-specific parameters, the data-free variant attains the highest mean OP. Second, the replay variant matches or exceeds SAE-FD, the other activation-buffer method, with the highest or second-highest retention of all methods. The architectural reference O-LoRA attains a higher OP than the data-free variant in all three settings, and the highest OP overall on TRACE with Gemma, at the cost of parameters that grow with every task.

\emph{TRACE.} Among the methods that store no previous-task data (\cref{tab:trace}), the data-free variant attains OP~$=0.539\pm0.015$: $+0.160$ over unprotected fine-tuning, $+0.015$ over EWC (within one standard deviation), $+0.030$ over SI, $+0.044$ over MAS, and $+0.118$ over ELLA. Without storing any examples, it also lies between 2\% and 10\% example replay, within one standard deviation of both. The replay variant attains $0.559\pm0.009$, within one standard deviation of SAE-FD ($0.565\pm0.025$), which stores the same 1.4~GB of hidden states per task, and the strongest retention of all methods (BWT~$=-0.012$, ARA~$=0.970$). O-LoRA attains the highest OP ($0.627$) but adds adapter rank with every task; we treat it as an architectural reference.

\begin{figure}[t]
  \centering
  \includegraphics[width=\columnwidth]{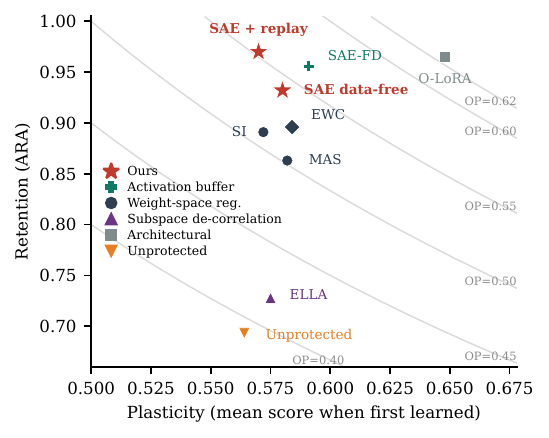}
  \caption{Stability--plasticity tradeoff on TRACE (Gemma-2 9B-it, three-seed means). Gray curves are iso-OP contours (OP $\approx$ ARA $\times$ plasticity). The weight-space regularizers share the plasticity of unprotected fine-tuning and differ mainly in retention; both SAE variants lie above them.}
  \label{fig:stability_plasticity}
\end{figure}

\Cref{fig:stability_plasticity} places each method in the stability--plasticity plane. The weight-space regularizers keep plasticity close to that of unprotected fine-tuning ($0.57$--$0.59$) and differ in retention. EWC and SI reduce the forgetting of unprotected fine-tuning by approximately two thirds (ARA near $0.90$ versus $0.69$), MAS by approximately one half (ARA~$=0.863$), whereas ELLA forgets more than twice as much as EWC (BWT~$=-0.154$ versus $-0.060$), although MAS and ELLA use the same $\lambda$ as EWC on importance scores rescaled to the mean magnitude of EWC's. ELLA and MAS are also the least stable methods ($\sigma=0.052$ and $0.038$): across seeds, their plasticity is nearly constant, whereas their ARA varies by up to $0.23$. At the same plasticity as EWC and SI ($0.580$), the data-free variant achieves higher retention (ARA~$=0.932$ versus $0.896$; \cref{sec:framework}).

\emph{MedCL.} The same pattern holds within the biomedical domain (\cref{tab:medcl}). With the settings selected on TRACE, the replay variant attains the highest OP of all methods, $0.620\pm0.009$, exceeding SAE-FD by $0.101$ and O-LoRA by $0.111$, and the strongest retention (ARA~$=0.926$, BWT~$=-0.042$) at the plasticity of unprotected fine-tuning. O-LoRA retains the second most (ARA~$=0.863$) but has the lowest plasticity in the table. Among the regularization methods that store no data, the data-free variant again attains the highest OP ($0.501\pm0.023$, against $0.477$ for EWC and $0.466$ for MAS) and the least forgetting (BWT~$=-0.134$ versus $-0.176$ and $-0.174$). As on TRACE, the regularizers reach similar plasticity and differ mainly in retention. Among the methods without a buffer, the gaps are smaller than on TRACE, plausibly because the more diverse task sequence of TRACE causes stronger interference.

\emph{Second model family.} On Llama-3.1-8B-Instruct (\cref{tab:trace_llama}), the replay variant ($0.601\pm0.004$), O-LoRA ($0.602\pm0.005$), and SAE-FD ($0.597\pm0.008$) are within one standard deviation of one another. The two activation-buffer methods retain nearly all acquired knowledge (ARA~$\approx0.98$), and ours is slightly more plastic at approximately one third of the per-step cost of SAE-FD (\cref{sec:cost}); O-LoRA retains less but is somewhat more plastic. The data-free variant ($0.575\pm0.006$) again attains the highest OP among the methods that store no data. Llama forgets less than Gemma (ARA of unprotected fine-tuning $0.92$ versus $0.69$), so the methods are more tightly clustered. With its validation-selected $\lambda$ ($10^2$, as on Gemma), EWC retains well (ARA~$=0.973$) but has the lowest plasticity in the table and improves OP over unprotected fine-tuning by only $0.008$; the replay variant matches this retention, and both SAE variants keep plasticity at or above the unprotected level.

\subsection{Computational Cost}
\label{sec:cost}

We profile every method in a controlled micro-benchmark: same model, data, LoRA configuration, and effective batch size, on a single dedicated GPU, for 120 optimizer steps after warmup (\cref{tab:overhead}). Regularizers that act from the second task on are profiled with a reference (weight anchor or stored hidden states) computed on the same task, since the penalty cost does not depend on which task the reference encodes; ELLA and O-LoRA are profiled as two-task chains.

\begin{table}[t]
\centering
\caption{Computational overhead (Gemma-2 9B-it, LoRA $r{=}256$, single A800, effective batch 8). Weight-space anchors are held on GPU. $^{a}$Anchor CPU-offloaded. $^{b}$Profiled with $r{=}16$ per-task adapters (half the $r{=}32$ of the main runs), so memory is not directly comparable. $^{c}$Per-task adapters merged into the base weights for deployment.}
\label{tab:overhead}
\footnotesize
\setlength{\tabcolsep}{2.6pt}
\begin{tabular}{@{}lccccc@{}}
\toprule
Method & Precomp. & s/step & GPU mem. & Store/task & Infer. \\
\midrule
Unprotected & --- & 1.70 & 30.0 GB & --- & none \\
\textbf{SAE data-free} & 37 s & 1.88 & \textbf{31.7 GB} & \textbf{197 KB} & none \\
\textbf{SAE + replay} & 97 s & 2.26 & 33.8 GB & 1.4 GB & none \\
EWC & 270 s & 1.80 & 37.3 GB & 6.5 GB & none \\
MAS & 289 s & 1.91 & 37.3 GB & 6.5 GB & none \\
SI & online & 1.89 & 47.7 GB & 6.5 GB & none \\
ELLA & 62 s & 3.33$^{a}$ & $\sim$26 GB & 16 GB & none \\
O-LoRA & 2 s & 1.78 & 21.4 GB$^{b}$ & 0.45 GB & none$^{c}$ \\
SAE-FD & 40 s & 6.54 & 35.2 GB & 1.4 GB & none \\
\bottomrule
\end{tabular}
\end{table}

Per-step cost is similar for most GPU-resident methods: EWC, MAS, SI, O-LoRA, and the data-free variant run at 1.78--1.91 s/step against 1.70 for unprotected training. The replay variant costs 2.26 s/step ($1.3\times$), since its additional forward pass covers only four stored sequences, at one layer; SAE-FD's dense distillation costs $3.8\times$. This parity in step time conceals a memory asymmetry: a weight-space penalty needs a reference copy of the trainable parameters and a per-parameter importance vector at every step, approximately 7~GB at $r{=}256$, or 17~GB for SI. Offloading it to the CPU slows EWC and MAS to approximately 5.4 s/step and SI to 9.2 s/step. The data-free variant's per-task state is a 197~KB mask (16,384 fp32 values at each of three layers), which is used only while that task is trained, and its only sizable object is the fixed SAE dictionary (352M parameters), shared across tasks and used only during training. No method adds inference cost: every method except O-LoRA deploys as a single LoRA adapter, and O-LoRA's per-task adapters can be merged into the base weights, although they accumulate in storage. Because a weight-space anchor scales with the number of trainable parameters while the mask scales only with the dictionary width, the gap widens with model size; the closed-form storage model, its validation against these measurements, and projections to 27B parameters are given in Appendix~\ref{app:cost}.

\subsection{Hyperparameter Sensitivity}
\label{sec:sweep}

\begin{figure}[t]
  \centering
  \includegraphics[width=\columnwidth]{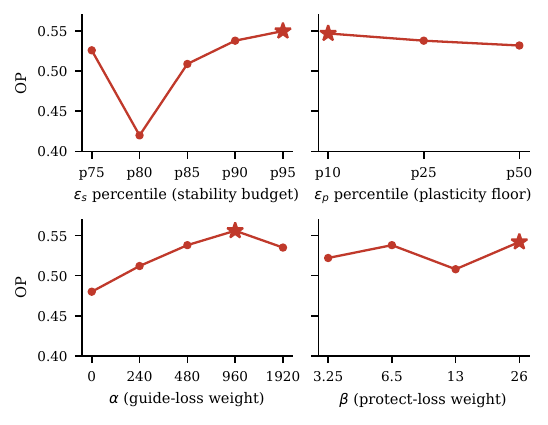}
  \caption{Hyperparameter sensitivity on TRACE (Gemma-2 9B-it, test split, single seed). One hyperparameter is varied per panel around the reference configuration; stars mark the per-axis optima.}
  \label{fig:sweep}
\end{figure}

We vary one hyperparameter at a time around the reference configuration ($\alpha{=}480$, $\beta{=}6.5$, $\varepsilon_s$ at the 90th and $\varepsilon_p$ at the 25th percentile of first-task drift) on the TRACE test split with a single seed (\cref{fig:sweep}). The reported configuration differs only in $\alpha{=}960$, which was selected on validation; the test-split results here characterize sensitivity and were not used for selection. The guide-loss weight $\alpha$ has the largest effect: OP rises from $0.480$ with the floor disabled to $0.556$ at $\alpha{=}960$ and decreases only to $0.535$ at $\alpha{=}1920$, a broad plateau. Performance is robust to loosening the stability budget $\varepsilon_s$ (OP $0.538$--$0.550$ for the 90th--95th percentiles) but degrades when the budget is tightened; at the 80th percentile, training is unstable (OP~$0.420$, reproduced on two seeds), although the neighboring settings are not. Performance is comparatively insensitive to $\varepsilon_p$ and $\beta$: OP varies by $0.015$ and $0.034$ across their ranges. Because the per-axis optima do not combine additively, all hyperparameters other than $\alpha$ are kept at their reference values. Around the reference configuration, moving any single hyperparameter by one step (halving or doubling $\alpha$ or $\beta$, or one percentile step for $\varepsilon_s$ or $\varepsilon_p$) changes OP by at most $0.03$. This is about twice the seed SD of the method ($0.015$, \cref{tab:trace}), and the non-monotone $\beta$ curve ($0.522$, $0.538$, $0.508$, $0.542$) is consistent with much of this variation being seed noise. The exceptions are disabling the guide loss ($\alpha{=}0$) and tightening $\varepsilon_s$ to the 80th percentile, discussed above.

\subsection{Ablation of the Loss Terms}
\label{sec:factorial}

\begin{table}[t]
\centering
\caption{Ablation of the loss terms on TRACE (Gemma-2 9B-it, OP, single seed). Active terms use their reference weights; the cell with both terms off is unprotected fine-tuning.}
\label{tab:factorial}
\footnotesize
\begin{tabular}{@{}lcc@{}}
\toprule
 & Guide loss on & Guide loss off \\
\midrule
Protect loss on  & \textbf{0.538} & 0.480 \\
Protect loss off & 0.385 & 0.351 \\
\bottomrule
\end{tabular}
\end{table}

A $2\times2$ factorial design switches the protect loss and the guide loss on and off independently (\cref{tab:factorial}). The protect loss alone raises OP from $0.351$ to $0.480$, whereas the guide loss alone raises it only to $0.385$ and leaves most of the forgetting (BWT~$=-0.181$, against $-0.054$ for the full method). Together, the two terms reach $0.538$, and the joint gain ($+0.187$) exceeds the sum of the individual gains ($+0.163$). This is consistent with the roles of the two terms in \cref{sec:formulation}: the protect loss is the stabilizer, whereas the guide loss, which constrains only the features in the plasticity set, contributes little on its own. Once the remaining features are preserved, it increases both plasticity ($0.577\to0.592$) and retention (ARA $0.845\to0.908$), consistent with directing the update toward the features of the task.

\section{Why Feature-Space Protection Is More Selective}
\label{sec:framework}

\Cref{sec:experiments} shows that, at similar plasticity, protection in the feature basis retains more than protection in the weight basis. We now test the account given in \cref{sec:intro}: that weight-space protection is non-selective because individual weights are polysemantic. Each MLP neuron adds its activation, multiplied by a fixed vector, its \emph{output direction} (the corresponding column of the MLP output projection), to the residual stream. We first measured the entanglement directly in Gemma-2 9B by decomposing each neuron's output direction over the SAE dictionary: the median neuron involves 500--1350 distinct SAE features, depending on the layer. This creates a granularity mismatch: when EWC protects weights for task~1, it also constrains the many features that pass through those weights, and if task~2 requires one of them to change, the penalty opposes that change. This account yields two testable predictions. Task-relevant units should be more linearly separable in the SAE basis than in the neuron basis, and protecting one task in weight space should constrain other tasks' features more than protection in the SAE basis does. Both tests support this account, although neither excludes other factors.

\subsection{Task Separability}
\label{sec:separability}

If a coordinate system supports concept-level protection, units relevant to different tasks should occupy distinguishable regions of it. For a fair comparison, we represent SAE features and MLP neurons as directions in the same $D$-dimensional residual stream. For each task pair we take the top-5\% units per task in each basis: SAE features with the highest activation relevance, represented by decoder vectors, and MLP neurons with the highest task-specific Fisher importance (aggregated over each neuron's parameters), represented by their output directions. A logistic classifier under 5-fold cross-validation predicts each direction's task, and we report ROC-AUC (chance 0.50), concatenating SAE layers 9, 20, and 31 and all MLP layers. Per benchmark we use three pairs spanning cross-domain, within-domain, and intermediate similarity (on TRACE: Py150 vs 20Minuten, NumGLUE-cm vs NumGLUE-ds, C-STANCE vs FOMC).

\begin{table}[t]
\centering
\caption{Linear separability of task-relevant units (5-fold CV ROC-AUC, chance 0.50), pooled over three task pairs per benchmark.}
\label{tab:separability}
\footnotesize
\begin{tabular}{@{}lcc@{}}
\toprule
Coordinate system & MedCL & TRACE \\
\midrule
SAE feature directions & \textbf{0.866} & \textbf{0.882} \\
MLP neuron directions & 0.502 & 0.498 \\
\bottomrule
\end{tabular}
\end{table}

Task-relevant SAE features are clearly separable (AUC $0.882$ on TRACE, $0.866$ on MedCL), whereas Fisher-important neurons are separable only at chance level ($0.498$, $0.502$; \cref{tab:separability}): polysemantic neurons participate in both tasks' computations, and their output directions carry little task-discriminative structure. The contrast holds both on the all-biomedical MedCL suite, where tasks share vocabulary, and on the diverse TRACE suite, which suggests that it reflects a property of the basis rather than domain dissimilarity. Per layer, the feature-basis AUC increases with depth (mean $0.67$ at layer~9 to $0.75$ at layer~31) and is higher for cross-domain pairs; per-layer projections are shown in \textcolor{figref}{Fig.~\ref{fig:sae_geometry}} in the appendix.

\subsection{Collateral Constraint}
\label{sec:collateral}

The second test quantifies the extent to which protecting the units of one task inadvertently constrains the features of another. When EWC protects a set of neurons for task~A, their output directions span a subspace of the residual stream, and any SAE feature whose decoder direction overlaps it is effectively constrained. We define a feature's \emph{exposure} as the fraction of its decoder-direction magnitude that projects onto task~A's protected subspace (top-5\% neurons by Fisher importance). Exposure of task~A's own relevant features is the \emph{intended} exposure; exposure of task~B's is the \emph{collateral} exposure; their ratio measures selectivity, with 1.0 meaning fully non-selective. For feature-space protection we compute the analogous ratio in the SAE basis, using the overlap with task~A's and task~B's top-5\% features. We evaluate all task pairs and all three layers on both benchmarks.

\begin{table}[t]
\centering
\caption{Collateral constraint: ratio of cross-task (collateral) to same-task (intended) exposure. Near 1.0 is non-selective; lower is more selective.}
\label{tab:collateral}
\footnotesize
\setlength{\tabcolsep}{4pt}
\begin{tabular}{@{}lccc@{}}
\toprule
Benchmark & Weight prot. & Feature prot. & Gap \\
\midrule
MedCL (biomedical) & 0.96 & 0.61 & 0.35 \\
TRACE (diverse) & 0.91 & 0.43 & 0.48 \\
\bottomrule
\end{tabular}
\end{table}

Under weight-space protection the ratio is 0.96 on MedCL and 0.91 on TRACE (\cref{tab:collateral}): even with task-specific Fisher importance, protecting task~A's neurons constrains task~B's features at 91--96\% of the rate of task~A's own, because features are constrained according to whether they pass through the protected neurons, not according to which task they serve. Under feature-space protection, the ratio decreases to 0.61 and 0.43. The gap is wider on TRACE, where tasks span different domains and their features occupy more distinct regions of the dictionary; MedCL tasks share biomedical concepts, so even concept-level protection exhibits greater overlap.

\subsection{Consequence for Weight-Space Regularization}
\label{sec:consequence}

These measurements help explain the behavior of the weight-space regularizers in \cref{sec:experiments}. EWC's $\lambda$ grid (Table~\ref{tab:val_ewc}) shows both ends of the tradeoff: at $\lambda{=}10^4$, ARA reaches $0.991$ but plasticity drops to $0.412$ (one seed), far below unprotected fine-tuning; at the selected $\lambda{=}10^2$, plasticity recovers to $0.584$ but ARA falls to $0.896$. If protecting task~$k$'s weights constrains the next task's features nearly as much as task~$k$'s own, then any $\lambda$ strong enough to prevent forgetting suppresses plasticity, and any $\lambda$ weak enough to permit learning no longer prevents forgetting. With a collateral ratio of 0.43--0.61, feature-space protection leaves many incoming-task features less constrained, consistent with the data-free variant combining ARA~$=0.932$ with plasticity $0.580$ on TRACE.

\section{Conclusion}

We introduced SAE-guided activation regularization for continual learning in LLMs, which constrains feature drift in the basis of a frozen sparse autoencoder through a loss derived as a regularized Lagrangian relaxation of explicit stability and plasticity constraints, with EWC as a special case. On TRACE with two model families and on MedCL, the data-free variant attains the highest mean overall performance among methods that store no previous-task data and add no task-specific parameters, with a per-task state of 197~KB, and the replay variant matches or exceeds the strongest activation-buffer method at approximately one third of its per-step cost. The separability and collateral-constraint analyses show that weight-space protection is nearly non-selective across tasks, whereas protection in the SAE basis is markedly more selective: the effectiveness of regularization depends not only on the penalty strength but also on the representation space in which the constraint is imposed.

\clearpage
\appendices

\section{Relation to EWC}
\label{app:ewc}

EWC is a special case of Eq.~(\ref{eq:hinge}), obtained by five choices: (i)~the coordinates are the weights ($f_i=\theta_i$, with no token dependence); (ii)~the reference is the post-task checkpoint $\boldsymbol{\theta}^\ast$; (iii)~the stability score is the diagonal Fisher information ($p_i=\mathcal{F}_i$, which may exceed~1), with $\beta/F=\lambda$; (iv)~the budget is zero ($\varepsilon_s=0$); and (v)~there is no plasticity constraint ($\alpha=0$). The contribution of coordinate $i$ to $\beta\,\mathcal{L}_{\mathrm{protect}}$ then becomes
\begin{equation}
  \lambda\,\mathcal{F}_i\max\bigl(0,|\theta_i-\theta_i^\ast|\bigr)^2=\lambda\,\mathcal{F}_i(\theta_i-\theta_i^\ast)^2,
  \label{eq:ewc}
\end{equation}
which is the EWC penalty, term by term.

Our method departs from EWC in choices (i), (iv), and (v), and each departure removes a specific limitation. In the weight basis, each coordinate is entangled across concepts; a zero budget penalizes every change, including those the current task requires; and without a plasticity constraint, forgetting can be reduced simply by learning less. Each departure is evaluated empirically: the basis by the separability and collateral analyses (Sections~\ref{sec:separability} and~\ref{sec:collateral}), the nonzero budget by the sweep over $\varepsilon_s$ (Section~\ref{sec:sweep}), with EWC's penalty-strength sweep (Section~\ref{sec:consequence}) showing the tradeoff left by a zero budget, and the floor by the term ablation (Section~\ref{sec:factorial}).

\section{Hyperparameter Selection and Settings}
\label{app:hparams}

\emph{Validation split.} TRACE ships a development split alongside its test split. We use the development split for all hyperparameter selection and report every result on the test split. FOMC has no separate development set, so it is excluded from all validation scores.

\emph{Reference values.} The loss has four hyperparameters: the stability budget $\varepsilon_s$, the plasticity floor $\varepsilon_p$, and the ceiling weight $\beta$ and the floor weight $\alpha$. We first fix a reference configuration from the drift observed under unregularized fine-tuning on the first task: $\varepsilon_s$ is set per layer to the 90th percentile of drift among features in the stability set ($m_{t,i}<0.1$), $\varepsilon_p$ to the 25th percentile among the 64 highest-scoring features in the plasticity set, and $\beta$ and $\alpha$ so that each loss term equals 5\% of the cross-entropy at the unregularized drift level. For Gemma on TRACE this gives $\alpha{=}480$, $\beta{=}6.5$. We then select $\alpha$, the most sensitive parameter (Section~\ref{sec:sweep}), from $\{240,480,960,1920\}$ on the validation split with the other three at the reference (\cref{tab:val_alpha}). $\alpha{=}960$ is selected and used for every Gemma TRACE result, in both variants. Llama uses a different SAE, a JumpReLU SAE trained at layer~20, so its hyperparameters are re-derived in the same way rather than carried over: the reference rule gives $\alpha{=}28$ and $\beta{=}29$, and one weight is then selected on the TRACE validation split, here $\beta$ from $\{29,100,1000,3000\}$, which gives $\beta{=}1000$. Both variants use $\alpha{=}28$ at layer~20 on Llama.

\begin{table}[t]
\centering
\caption{Validation OP on TRACE (Gemma-2 9B-it, seed 42) for the $\alpha$ grid; other hyperparameters at the reference values.}
\label{tab:val_alpha}
\footnotesize
\begin{tabular}{@{}lcccc@{}}
\toprule
$\alpha$ & 240 & 480 & \textbf{960} & 1920 \\
\midrule
Validation OP & 0.498 & 0.520 & \textbf{0.549} & 0.517 \\
\bottomrule
\end{tabular}
\end{table}

\Cref{tab:val_sweep} gives validation scores for the full one-factor-at-a-time sweep of Section~\ref{sec:sweep}. The validation split ranks the $\alpha$ axis exactly as the test split does. On the other three axes the validation optimum differs from the reference by at most $0.018$; we do not act on these, because the per-axis optima do not compose.

\begin{table}[t]
\centering
\caption{Validation OP (seed 42) for the one-factor-at-a-time sweep around the reference ($\alpha{=}480$, $\beta{=}6.5$, $\varepsilon_s$ at p90, $\varepsilon_p$ at p25).}
\label{tab:val_sweep}
\footnotesize
\setlength{\tabcolsep}{4pt}
\begin{tabular}{@{}lllll@{}}
\toprule
Axis & \multicolumn{4}{l}{Validation OP at each setting} \\
\midrule
$\alpha$ & 240: 0.498 & 480: 0.520 & 960: 0.549 & 1920: 0.517 \\
$\varepsilon_s$ & p85: 0.489 & p90: 0.520 & p95: 0.529 & \\
$\beta$ & 3.25: 0.498 & 6.5: 0.520 & 26: 0.533 & \\
$\varepsilon_p$ & p10: 0.538 & p25: 0.520 & p50: 0.524 & \\
\bottomrule
\end{tabular}
\end{table}

On Llama, the reference $\beta$ leaves the protection term much weaker in training than on Gemma: its median ratio to the cross-entropy on the last task is about 0.16, against 3.6 on Gemma. We therefore select $\beta$ rather than $\alpha$ on Llama (\cref{tab:val_llama}).

\begin{table}[t]
\centering
\caption{Validation OP on TRACE (Llama-3.1-8B-Instruct, one seed) for the $\beta$ grid; $\alpha{=}28$ and $\varepsilon_s$, $\varepsilon_p$ at the reference values.}
\label{tab:val_llama}
\footnotesize
\begin{tabular}{@{}lcccc@{}}
\toprule
$\beta$ & 29 & 100 & \textbf{1000} & 3000 \\
\midrule
Validation OP & 0.556 & 0.561 & \textbf{0.568} & 0.554 \\
\bottomrule
\end{tabular}
\end{table}

\emph{Baseline selection.} EWC's weight is selected on the same split from $\lambda\in\{10,10^2,10^3,10^4\}$, with one seed for every value (\cref{tab:val_ewc}). $\lambda{=}10^2$ attains the highest validation OP and is used for all Gemma EWC results. A weaker penalty forgets more without learning new tasks better, and stronger penalties over-regularize: at $\lambda{=}10^4$ the model retains almost everything it has learned but learns little. The selected $\lambda{=}10^2$ is used unchanged for Gemma on MedCL. On Llama, both EWC and our method are re-tuned: $\lambda$ is re-selected in the same way from $\{10,10^2,10^3,3{\times}10^3\}$ (\cref{tab:val_ewc_llama}), which again selects $\lambda{=}10^2$.

\begin{table}[t]
\centering
\caption{Validation OP on TRACE (Gemma-2 9B-it, one seed) for the EWC $\lambda$ grid.}
\label{tab:val_ewc}
\footnotesize
\begin{tabular}{@{}lcccc@{}}
\toprule
$\lambda$ & $10$ & \textbf{$10^{2}$} & $10^{3}$ & $10^{4}$ \\
\midrule
Validation OP & 0.512 & \textbf{0.554} & 0.498 & 0.360 \\
\bottomrule
\end{tabular}
\end{table}

\begin{table}[t]
\centering
\caption{Validation OP on TRACE (Llama-3.1-8B-Instruct, one seed) for the EWC $\lambda$ grid.}
\label{tab:val_ewc_llama}
\footnotesize
\begin{tabular}{@{}lcccc@{}}
\toprule
$\lambda$ & $10$ & \textbf{$10^{2}$} & $10^{3}$ & $3{\times}10^{3}$ \\
\midrule
Validation OP & 0.504 & \textbf{0.526} & 0.483 & 0.441 \\
\bottomrule
\end{tabular}
\end{table}

\emph{Settings of all methods.} \Cref{tab:hparams} lists the method-specific settings of every Gemma TRACE run. MedCL uses the Gemma TRACE settings unchanged. On Llama, both variants use the custom layer-20 SAE with $\alpha{=}28$, $\varepsilon_s{=}0.15$ and $\varepsilon_p{=}0.032$; the data-free variant uses $\beta{=}1000$, and the replay variant uses the Gemma buffer settings at layer~20. Shared settings (LoRA rank and scaling, optimizer, learning rate, effective batch size, epochs, sequence length) are identical across methods on a given benchmark. Only EWC's $\lambda$ is selected among the baselines. MAS and ELLA reuse the selected $\lambda{=}10^2$ without further tuning; their importance scores are rescaled so that their mean magnitude matches EWC's diagonal Fisher on the same benchmark, so that the same weight acts on a comparable scale. SI uses its standard settings, which were not tuned. O-LoRA and SAE-FD use the method-specific settings of their original papers, with adapter rank, batch size, and learning rate matched to all other methods. All settings selected on Gemma TRACE are used unchanged on MedCL.

\begin{table*}[t]
\centering
\caption{Method-specific hyperparameters (TRACE, Gemma-2 9B-it).}
\label{tab:hparams}
\footnotesize
\begin{tabular}{@{}l p{0.78\textwidth}@{}}
\toprule
Method & Settings \\
\midrule
SAE data-free (ours) & $\alpha{=}960$ (validation-selected), $\beta{=}6.5$; $\varepsilon_s$ at p90 and $\varepsilon_p$ at p25 of first-task drift, per layer (L9, L20, L31) \\
SAE + replay (ours) & $\alpha{=}960$ at L31 only; buffer protect weight $\beta{=}20$ with per-feature budget $\varepsilon_{s,i}{=}\kappa\sigma_i$, $\kappa{=}1.0$, at L31; 200 stored sequences per task, 4 per step; protect loss capped at $0.5\times$ the task loss \\
EWC & $\lambda{=}10^{2}$ (validation-selected on Gemma TRACE and used unchanged on MedCL; re-selected on Llama, \cref{tab:val_ewc_llama}, which again gives $10^{2}$); cumulative diagonal Fisher over 256 batches per task \\
SI & $\lambda{=}1$, $\xi{=}10^{-3}$ (standard settings, not selected); online path integral, merged cumulatively \\
MAS & $\lambda{=}10^{2}$ (EWC's selected value, not tuned separately; TRACE and MedCL); importance over 256 batches per task, rescaled to the mean magnitude of EWC's Fisher on the same benchmark \\
ELLA & $\lambda{=}10^{2}$ (EWC's selected value, not tuned separately); aggregated past update $\sum_{i<t}\Delta\mathbf{W}_i$, rescaled to the mean magnitude of EWC's Fisher \\
O-LoRA & per-task LoRA $r{=}32$, $\alpha_{\mathrm{LoRA}}{=}64$; orthogonality weight $0.5$ (same on both models) \\
SAE-FD & $\rho{=}0.15$, EMA $0.05$, $\lambda\in[0.2,10]$; per-task stored activations \\
Example replay & 2\% or 10\% of each previous task's training set; no other settings \\
\bottomrule
\end{tabular}
\end{table*}

\section{Storage Model and Scaling}
\label{app:cost}

\emph{CPU offloading.} Holding weight-space anchors on GPU increases peak memory by about 7~GB for EWC and MAS and 17~GB for SI (Table~\ref{tab:overhead}). Offloading them to CPU removes the memory cost but the per-step host-to-device transfer slows EWC to 5.42 s/step ($3.2\times$ unprotected training), MAS to 5.39 s/step, and SI to 9.16 s/step ($5.4\times$). Per-parameter methods thus pay either in device memory or in training time; the SAE dictionary (352M parameters, most of the $+1.7$~GB increment of the data-free variant) is fixed and shared across tasks.

\emph{Closed-form storage.} Let $N$ be the number of trainable parameters, $L$ the number of regularized layers, $F$ the dictionary width, $D$ the residual-stream dimension, $b$ the bytes per stored value, and $c$ the bytes of persistent state per trainable parameter. A weight-space penalty persists an anchor and an importance value for every trainable parameter per task, whereas feature-space regularization persists one mask per task plus a single dictionary shared by all tasks:	
\begin{gather}
  S_{\mathrm{weight}}=c\,N\quad(c=2b \text{ for EWC/MAS},\; c\approx4b \text{ for SI}), \label{eq:cost}\\
  S_{\mathrm{feat}}=b\,L\,F,\qquad S_{\mathrm{dict}}=2b\,L\,D\,F. \nonumber
\end{gather}
The per-task ratio $S_{\mathrm{weight}}/S_{\mathrm{feat}}=cN/(bLF)$ grows linearly with the number of trainable parameters, whereas the mask depends only on the dictionary. At our configuration ($N=864$M, $L{=}3$, $F{=}16{,}384$, fp32), \eqref{eq:cost} predicts a 6.9~GB anchor, close to the measured 6.5~GB.

\begin{table}[t]
\centering
\caption{Cost-model projections (LoRA $r{=}256$ on all linear projections unless noted; weight-space anchor at $c{=}8$ bytes). The 9B row is validated against measurement.}
\label{tab:scaling_cost}
\footnotesize
\setlength{\tabcolsep}{3pt}
\begin{tabular}{@{}lccc@{}}
\toprule
Configuration & Trainable & Weight anchor & Mask/task \\
\midrule
Gemma-2 2B & 0.33 B & 2.7 GB & 0.2 MB \\
Gemma-2 9B & 0.86 B & 6.9 GB (meas.\ 6.5) & 0.2 MB (meas.) \\
Gemma-2 27B & 1.83 B & 14.6 GB & 0.2 MB \\
Gemma-2 27B, full FT & 27 B & 216 GB & 0.2 MB \\
\bottomrule
\end{tabular}
\end{table}

At 27B (\cref{tab:scaling_cost}), keeping weight-space methods near baseline speed would require $+14.6$~GB on device for EWC and roughly $+29$~GB for SI; under full fine-tuning a 216~GB per-task anchor is impractical, while the mask is unchanged. The fixed dictionary cost $S_{\mathrm{dict}}$ grows with $D\cdot F$ and is about 0.9~GB in bf16 for three 16k-width layers at 27B. At 9B the dictionary (about 0.7~GB in bf16) is already smaller than a single EWC anchor, so the total state of feature-space regularization is below that of weight-space regularization from the first task onward. The replay variant stores 1.4~GB per task (about 11~GB over eight tasks), the same format and footprint as SAE-FD; ELLA retains 16~GB of reconstructed weight deltas, and O-LoRA's per-task adapters (0.45~GB as fp32 dumps; about 225~MB in bf16) accumulate.

\section{Per-Layer Feature Separability}
\label{app:separability}

\begin{figure*}[t]
  \centering
  \includegraphics[width=0.9\textwidth]{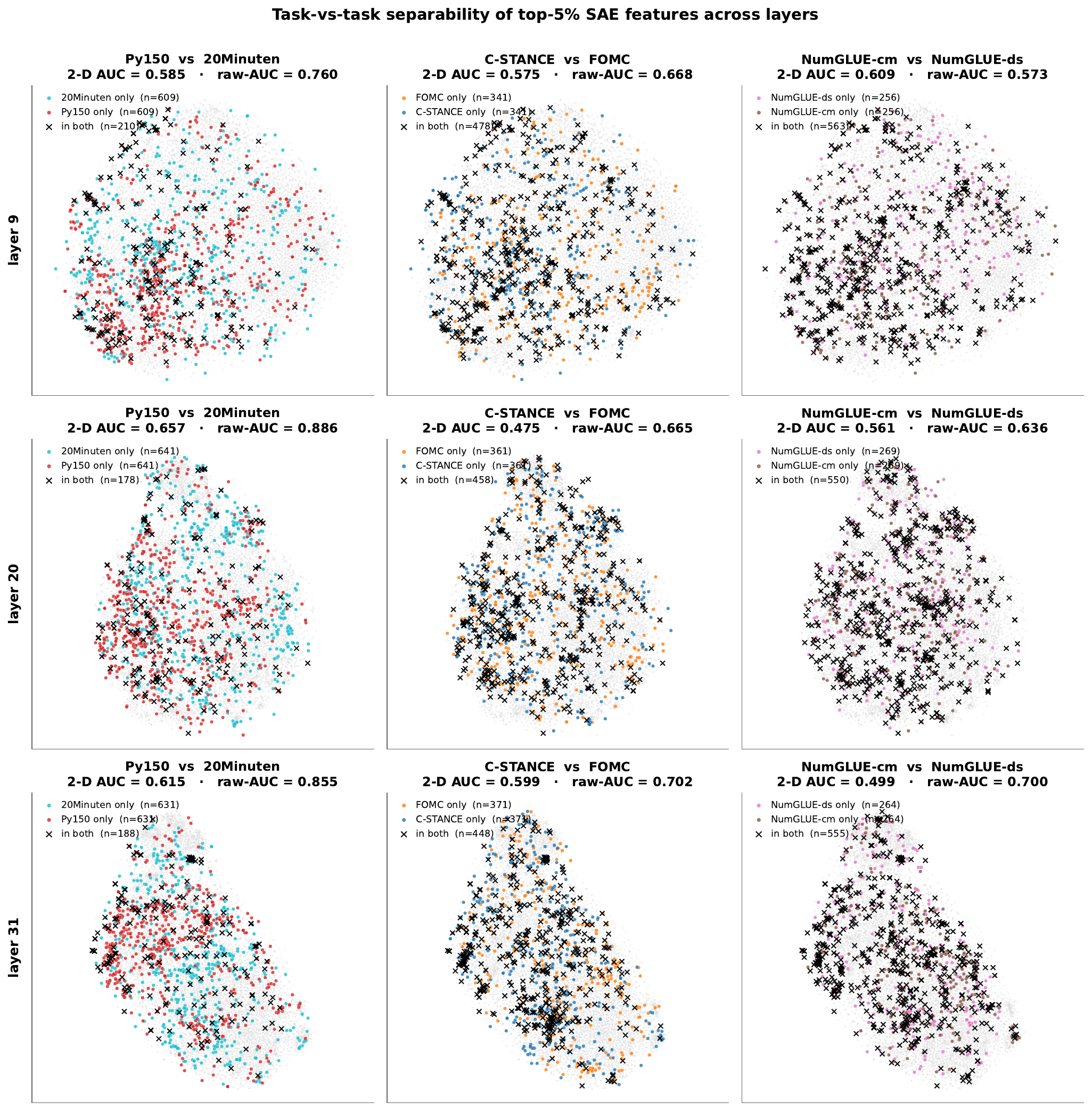}
  \caption{t-SNE projections of task-relevant SAE features for three TRACE task pairs at layers 9, 20, and 31. Each point is an SAE feature direction; color indicates task; $\times$ marks features relevant to both tasks. AUC values report linear separability of the decoder directions in the full $D$-dimensional residual stream ($D{=}3{,}584$).}
  \label{fig:sae_geometry}
\end{figure*}

\Cref{fig:sae_geometry} gives a qualitative view of the feature-basis separability of Table~\ref{tab:separability}. Per-layer AUC ranges from 0.57 for the most similar pair at the shallowest layer (NumGLUE-cm vs NumGLUE-ds, layer~9) to 0.89 for a cross-domain pair at mid-depth (Py150 vs 20Minuten, layer~20). Mean AUC rises from 0.67 at layer~9 to 0.75 at layer~31, consistent with deeper layers encoding more task-specific structure, and cross-domain pairs are more separable than within-domain pairs.

\section{Per-Task Results and Forgetting Dynamics}
\label{app:pertask}

\begin{figure}[t]
  \centering
  \includegraphics[width=\columnwidth]{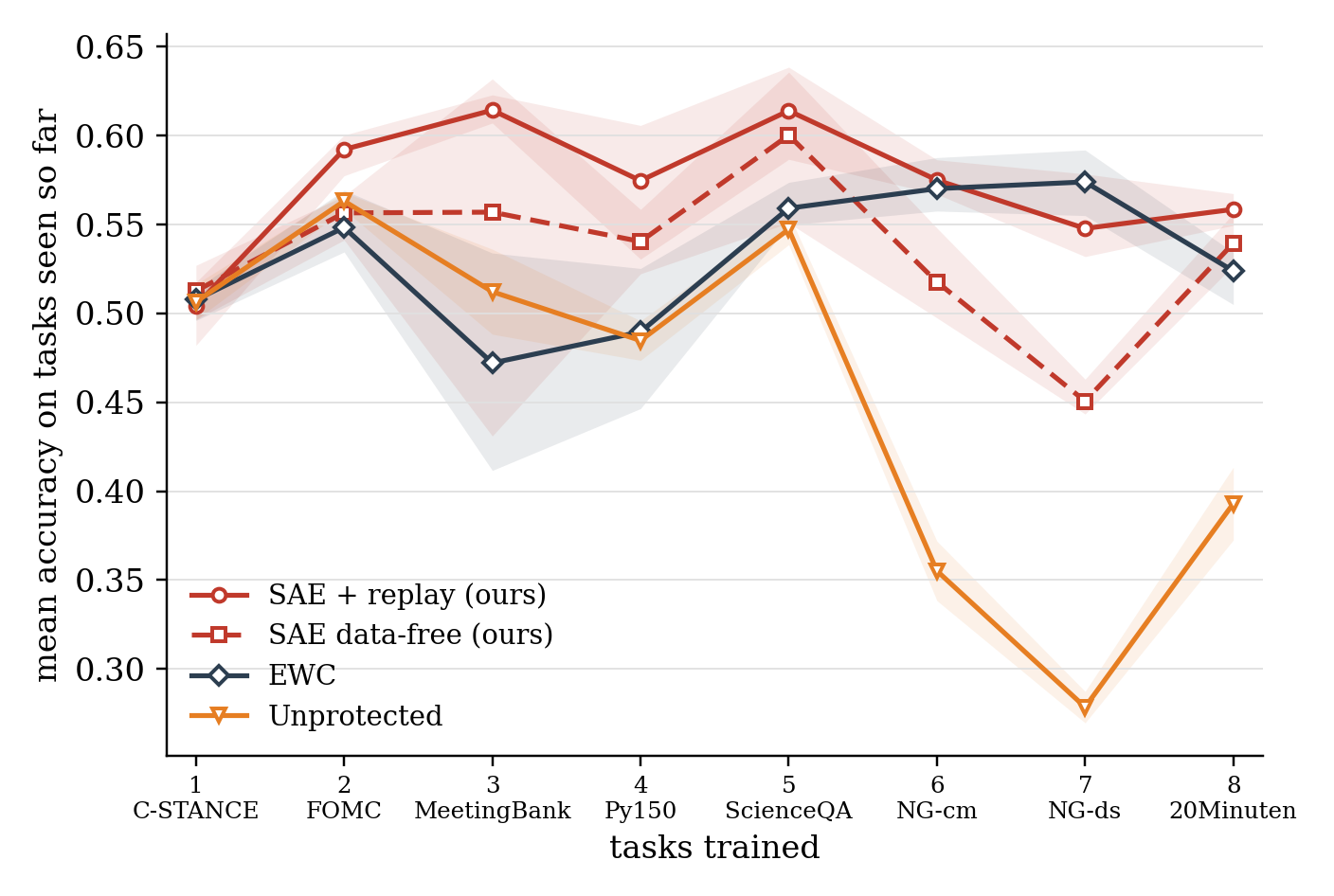}
  \caption{Forgetting on TRACE (Gemma-2 9B-it): mean score over the tasks seen so far, evaluated after each task. Lines are seed means; bands span the seed range (the unprotected band uses the two seeds with full evaluation matrices). The points at $k{=}8$ equal OP.}
  \label{fig:forgetting}
\end{figure}

\Cref{tab:pertask_trace,tab:pertask_medcl,tab:pertask_llama} break down the OP of every method in Tables~\ref{tab:trace}--\ref{tab:trace_llama} into per-task final scores. \Cref{fig:forgetting} shows when forgetting happens during the TRACE sequence, and \cref{tab:acc_matrix} gives the underlying checkpoint-by-task accuracy matrices for the replay variant and the unprotected baseline.

\begin{table*}[t]
\centering
\caption{Per-task final score on TRACE (Gemma-2 9B-it). Each cell is the score on that task after the full eight-task sequence, averaged over the three seeds of Table~\ref{tab:trace}; the Mean column equals OP in Table~\ref{tab:trace}.}
\label{tab:pertask_trace}
\footnotesize
\setlength{\tabcolsep}{3.2pt}
\begin{tabular}{@{}lccccccccc@{}}
\toprule
Method & C-STANCE & FOMC & MeetingBank & Py150 & ScienceQA & NG-cm & NG-ds & 20Minuten & Mean \\
\midrule
O-LoRA (architectural) & 0.554 & 0.733 & 0.594 & 0.567 & 0.924 & 0.704 & 0.754 & 0.190 & 0.627 \\
SAE-FD (activation buffer) & 0.455 & 0.642 & 0.641 & 0.547 & 0.885 & 0.449 & 0.715 & 0.190 & 0.565 \\
\textbf{SAE + replay (ours)} & 0.502 & 0.697 & 0.630 & 0.503 & 0.884 & 0.387 & 0.682 & 0.183 & 0.559 \\
Example replay 10\% & 0.473 & 0.652 & 0.630 & 0.498 & 0.910 & 0.321 & 0.661 & 0.191 & 0.542 \\
\textbf{SAE data-free (ours)} & 0.465 & 0.700 & 0.523 & 0.541 & 0.861 & 0.346 & 0.692 & 0.189 & 0.539 \\
Example replay 2\% & 0.481 & 0.634 & 0.613 & 0.504 & 0.891 & 0.296 & 0.667 & 0.190 & 0.534 \\
EWC & 0.342 & 0.391 & 0.662 & 0.564 & 0.879 & 0.580 & 0.578 & 0.194 & 0.524 \\
SI & 0.399 & 0.448 & 0.563 & 0.511 & 0.856 & 0.412 & 0.689 & 0.192 & 0.509 \\
MAS & 0.384 & 0.349 & 0.544 & 0.539 & 0.832 & 0.412 & 0.706 & 0.192 & 0.495 \\
ELLA & 0.179 & 0.177 & 0.284 & 0.505 & 0.817 & 0.481 & 0.730 & 0.194 & 0.421 \\
Unprotected & 0.158 & 0.169 & 0.470 & 0.368 & 0.752 & 0.280 & 0.640 & 0.194 & 0.379 \\
\bottomrule
\end{tabular}
\end{table*}
\begin{table*}[t]
\centering
\caption{Per-task final accuracy on MedCL (Gemma-2 9B-it, test split, log-probability evaluation), averaged over the three seeds of Table~\ref{tab:medcl}; the Mean column equals OP in Table~\ref{tab:medcl}.}
\label{tab:pertask_medcl}
\footnotesize
\setlength{\tabcolsep}{3.2pt}
\begin{tabular}{@{}lccccccccccc@{}}
\toprule
Method & pubmedqa & scifact & pubhealth & gad & chemprot & ddi & pubmed\_rct & litcovid & bioasq & druglib & Mean \\
\midrule
\textbf{SAE + replay (ours)} & 0.511 & 0.729 & 0.862 & 0.510 & 0.345 & 0.540 & 0.844 & 0.608 & 0.803 & 0.449 & 0.620 \\
SAE-FD (activation buffer) & 0.511 & 0.498 & 0.231 & 0.510 & 0.235 & 0.571 & 0.771 & 0.575 & 0.797 & 0.487 & 0.519 \\
O-LoRA (architectural) & 0.522 & 0.563 & 0.271 & 0.511 & 0.155 & 0.572 & 0.701 & 0.470 & 0.820 & 0.509 & 0.509 \\
\textbf{SAE data-free (ours)} & 0.582 & 0.498 & 0.225 & 0.510 & 0.217 & 0.541 & 0.671 & 0.492 & 0.800 & 0.472 & 0.501 \\
Unprotected & 0.531 & 0.498 & 0.183 & 0.510 & 0.181 & 0.523 & 0.675 & 0.475 & 0.781 & 0.461 & 0.482 \\
EWC & 0.538 & 0.498 & 0.221 & 0.510 & 0.173 & 0.525 & 0.620 & 0.461 & 0.803 & 0.418 & 0.477 \\
MAS & 0.540 & 0.499 & 0.197 & 0.510 & 0.177 & 0.482 & 0.723 & 0.406 & 0.745 & 0.380 & 0.466 \\
\bottomrule
\end{tabular}
\end{table*}
\begin{table*}[t]
\centering
\caption{Per-task final score on TRACE (Llama-3.1-8B-Instruct, seed 42). The Mean column is the seed-42 OP; for the rows reported as three-seed means in Table~\ref{tab:trace_llama} it differs slightly from the tabulated mean.}
\label{tab:pertask_llama}
\footnotesize
\setlength{\tabcolsep}{3.2pt}
\begin{tabular}{@{}lccccccccc@{}}
\toprule
Method & C-STANCE & FOMC & MeetingBank & Py150 & ScienceQA & NG-cm & NG-ds & 20Minuten & Mean \\
\midrule
O-LoRA (architectural) & 0.525 & 0.677 & 0.571 & 0.577 & 0.916 & 0.605 & 0.720 & 0.178 & 0.596 \\
\textbf{SAE + replay (ours)} & 0.540 & 0.712 & 0.600 & 0.561 & 0.911 & 0.580 & 0.735 & 0.180 & 0.602 \\
SAE-FD (activation buffer) & 0.554 & 0.692 & 0.601 & 0.568 & 0.920 & 0.531 & 0.655 & 0.177 & 0.587 \\
\textbf{SAE data-free (ours)} & 0.546 & 0.688 & 0.431 & 0.550 & 0.909 & 0.556 & 0.680 & 0.183 & 0.568 \\
EWC & 0.522 & 0.704 & 0.655 & 0.521 & 0.861 & 0.506 & 0.569 & 0.177 & 0.564 \\
Unprotected & 0.505 & 0.669 & 0.493 & 0.548 & 0.879 & 0.481 & 0.729 & 0.184 & 0.561 \\
\bottomrule
\end{tabular}
\end{table*}
\begin{table*}[t]
\centering
\caption{Task-score matrices on TRACE (Gemma-2 9B-it). Row $k$ gives the score on every task learned so far, evaluated at the checkpoint after training task $k$; the diagonal is each task's just-learned score and the last row its final score. Unprotected fine-tuning is shown at seed 43 because the legacy seed-42 evaluation stores no intermediate checkpoints.}
\label{tab:acc_matrix}
\footnotesize
\setlength{\tabcolsep}{5pt}
\begin{tabular}{@{}lcccccccc@{}}
\toprule
After task & C-STANCE & FOMC & MeetingBank & Py150 & ScienceQA & NG-cm & NG-ds & 20Minuten \\
\midrule
\multicolumn{9}{@{}l}{\textit{SAE + replay (ours), seed 42}} \\[1pt]
$k=1$ & 0.482 & -- & -- & -- & -- & -- & -- & -- \\
$k=2$ & 0.479 & 0.675 & -- & -- & -- & -- & -- & -- \\
$k=3$ & 0.489 & 0.688 & 0.662 & -- & -- & -- & -- & -- \\
$k=4$ & 0.476 & 0.675 & 0.651 & 0.548 & -- & -- & -- & -- \\
$k=5$ & 0.474 & 0.702 & 0.649 & 0.480 & 0.888 & -- & -- & -- \\
$k=6$ & 0.470 & 0.690 & 0.439 & 0.511 & 0.863 & 0.457 & -- & -- \\
$k=7$ & 0.463 & 0.692 & 0.472 & 0.505 & 0.818 & 0.444 & 0.655 & -- \\
$k=8$ & 0.481 & 0.708 & 0.630 & 0.553 & 0.898 & 0.407 & 0.677 & 0.185 \\
\midrule
\multicolumn{9}{@{}l}{\textit{Unprotected, seed 43}} \\[1pt]
$k=1$ & 0.496 & -- & -- & -- & -- & -- & -- & -- \\
$k=2$ & 0.432 & 0.704 & -- & -- & -- & -- & -- & -- \\
$k=3$ & 0.369 & 0.438 & 0.658 & -- & -- & -- & -- & -- \\
$k=4$ & 0.351 & 0.448 & 0.560 & 0.536 & -- & -- & -- & -- \\
$k=5$ & 0.450 & 0.462 & 0.536 & 0.333 & 0.913 & -- & -- & -- \\
$k=6$ & 0.312 & 0.456 & 0.065 & 0.318 & 0.545 & 0.333 & -- & -- \\
$k=7$ & 0.157 & 0.220 & 0.000 & 0.157 & 0.489 & 0.222 & 0.643 & -- \\
$k=8$ & 0.172 & 0.107 & 0.482 & 0.400 & 0.789 & 0.272 & 0.563 & 0.196 \\
\bottomrule
\end{tabular}
\end{table*}

\clearpage
\bibliographystyle{IEEEtran}
\bibliography{references}

\end{document}